\documentclass{article} 
\usepackage{iclr2027_conference,times}

\usepackage{amsmath,amsfonts,bm}

\def\eqref#1{equation~\ref{#1}}

\def\1{\bm{1}}

\DeclareMathAlphabet{\mathsfit}{\encodingdefault}{\sfdefault}{m}{sl}
\SetMathAlphabet{\mathsfit}{bold}{\encodingdefault}{\sfdefault}{bx}{n}

\usepackage[table]{xcolor}     
\definecolor{color3}{RGB}{240,240,250}
\definecolor{citeblue}{RGB}{0,0,128}
\definecolor{softgreen}{RGB}{0,150,0}
\usepackage{hyperref}         
\hypersetup{
    colorlinks = true,
    citecolor  = citeblue,     
    linkcolor  = red,           
    urlcolor   = blue,      
}
\usepackage{url}
\usepackage{graphicx}
\usepackage{wrapfig}
\usepackage{float}
\usepackage{booktabs,multirow}

\title{RATIO: Reasoning Analysis and Token-level Inference Optimization for Quantized Reasoning Models}

\author{%
\parbox[t]{\dimexpr\textwidth-2\tabcolsep\relax}{%
\centering
\textbf{Chengzhu Bao}\textsuperscript{1*}, \quad
\textbf{Xianglong Yan}\textsuperscript{1*}, \quad
\textbf{Tianao Zhang}\textsuperscript{1}, \\
\textbf{Jiaqi Chen}\textsuperscript{1}, \quad
\textbf{Shaoqiu Zhang}\textsuperscript{1}, \quad
\textbf{Yulun Zhang}\textsuperscript{1$\dagger$} \\
\normalfont\textsuperscript{1}Shanghai Jiao Tong University
}%
}

\iclrfinalcopy 
\begin{document}

\maketitle
\lhead{} 
\begingroup
\renewcommand{\thefootnote}{\fnsymbol{footnote}}
\footnotetext[1]{Equal contribution.}
\footnotetext[2]{Corresponding author: Yulun Zhang, yulun100@gmail.com.}
\endgroup

\vspace{-6mm}
\begin{abstract}
\vspace{-4mm}
Post-training quantization (PTQ) has become a widely adopted technique for reducing the memory footprint and inference cost of large language models (LLMs). However, recent studies reveal that when applied to reasoning models, PTQ not only degrades reasoning performance but also exacerbates overthinking, leading to longer reasoning trajectories. These issues may offset the efficiency gains expected from lower-precision inference. Existing approaches mainly rely on complex optimization procedures. More recent lightweight inference strategies instead use predefined overthinking markers, limiting their adaptability across quantized models. To address these issues, we propose \underline{R}easoning \underline{A}nalysis and \underline{T}oken-level \underline{I}nference \underline{O}ptimization (RATIO), a framework that identifies model-specific overthinking tokens and assigns each a tailored penalty. RATIO first introduces Quantization-aware Reasoning Behavior Analysis (QRBA) to identify overthinking tokens by analyzing discrepancies between full-precision and quantized models. It then adopts Token-Specific Penalty Determination (TSPD), which leverages full-precision guidance to derive token-specific penalties without additional training. Extensive experiments show that RATIO achieves a better accuracy-efficiency trade-off than existing token-level interventions. Specifically, RATIO achieves up to 9.8 points accuracy improvement and reduces chain-of-thought (CoT) length by up to 51.3\% compared with quantized baselines. The code will be available at \url{https://github.com/steven-bao1/RATIO}.
\end{abstract}

\setlength{\abovedisplayskip}{2pt}
\setlength{\belowdisplayskip}{2pt}
\setlength{\abovedisplayshortskip}{0pt}
\setlength{\belowdisplayshortskip}{2pt}

\vspace{-6mm}
\section{Introduction}
\begin{wrapfigure}[16]{r}{0.52\textwidth}
    \vspace{-3.8\baselineskip}
    \centering
    \includegraphics[width=\linewidth]{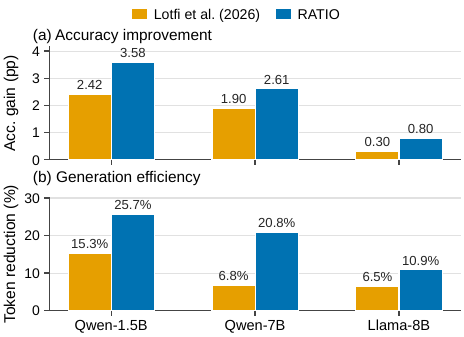}
    \vspace{-6mm}
    \caption{Performance comparison of different strategies across three quantized reasoning models. RATIO consistently achieves higher accuracy gains and larger CoT length reductions.}
    \vspace{-5mm}
    \label{fig:fig1}
\end{wrapfigure}
Large language models (LLMs) have achieved remarkable performance across a wide range of tasks, with increasing model size playing an important role in their progress. Therefore, modern LLMs~\citep{kimiteam2026kimik3,deepseekai2026deepseekv4,glm5} continue to grow to further improve performance. However, this growth increases memory requirements and inference costs, which hinder the practical deployment of LLMs. Post-training quantization (PTQ)~\citep{frantar_gptq_2023,lin_awq_2024,ashkboos_quarot_2024,yan2026d2quant} mitigates these costs by representing model parameters at lower precision, reducing memory usage and per-step decoding costs. These benefits have made PTQ a widely adopted approach to LLM deployment.

However, applying PTQ to reasoning models introduces new challenges, as large reasoning models (LRMs) rely on extended inference-time computation and generate long reasoning trajectories to solve complex tasks. First, low-bit quantization can introduce representation errors that degrade the final accuracy of reasoning models~\citep{liu2025quantizationhurts}. More importantly, recent studies~\citep{lotfi2026quantized,lian2026quantizationinflates} have found that quantization can exacerbate overthinking behaviors, such as hesitation and repeated verification. This can lead to longer chains of thought and greater token usage, potentially offsetting the inference efficiency gains expected from quantization.
\begin{figure}[t]
    \centering
    \begin{minipage}[t]{0.55\textwidth}
        \centering
        \includegraphics[width=\linewidth,height=1.70in,keepaspectratio]{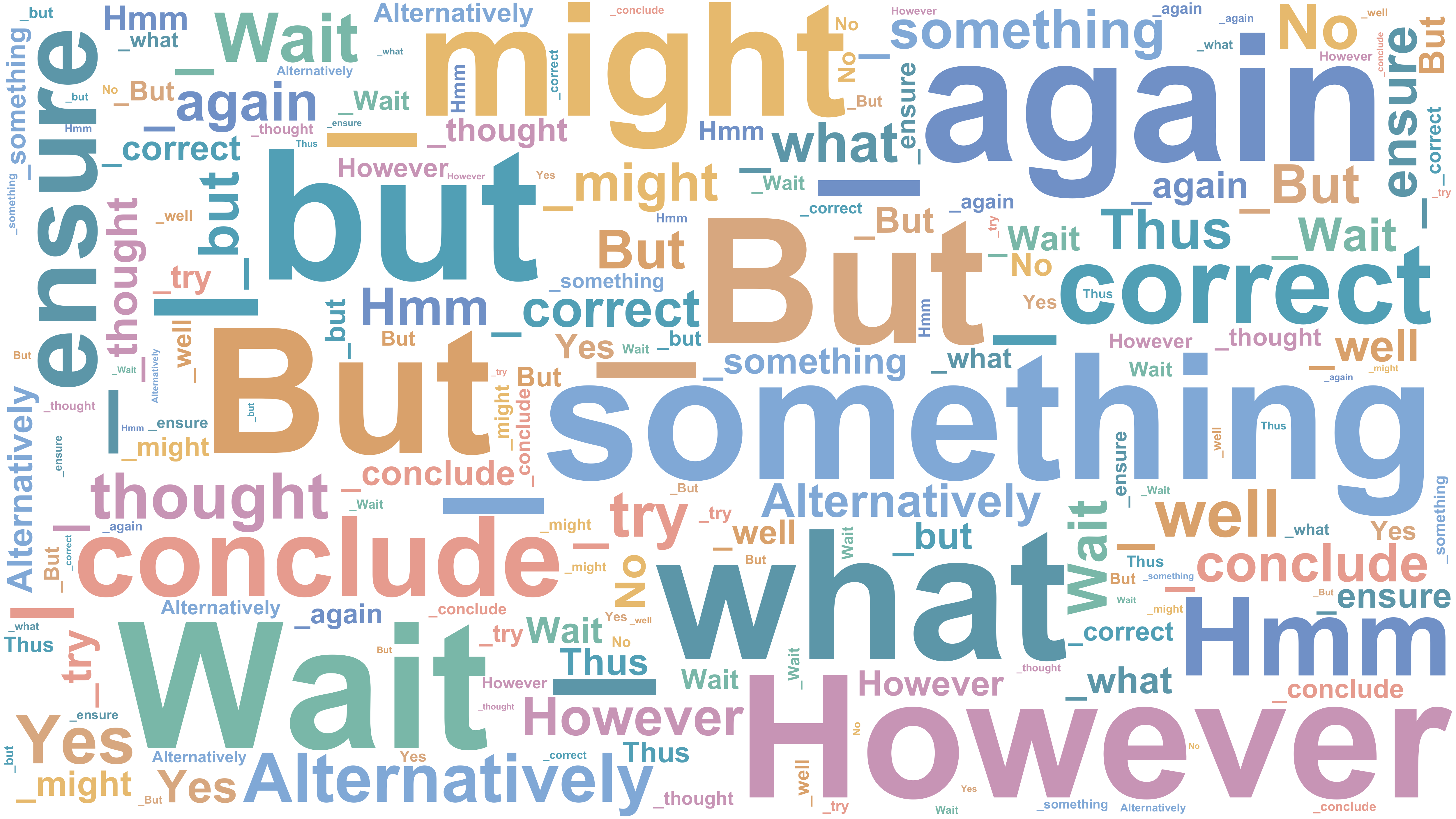}
        \par\smallskip
        {\small (a) Selected reasoning tokens\par}
    \end{minipage}\hfill
    \begin{minipage}[t]{0.43\textwidth}
        \centering
        \includegraphics[width=\linewidth,height=1.70in,keepaspectratio]{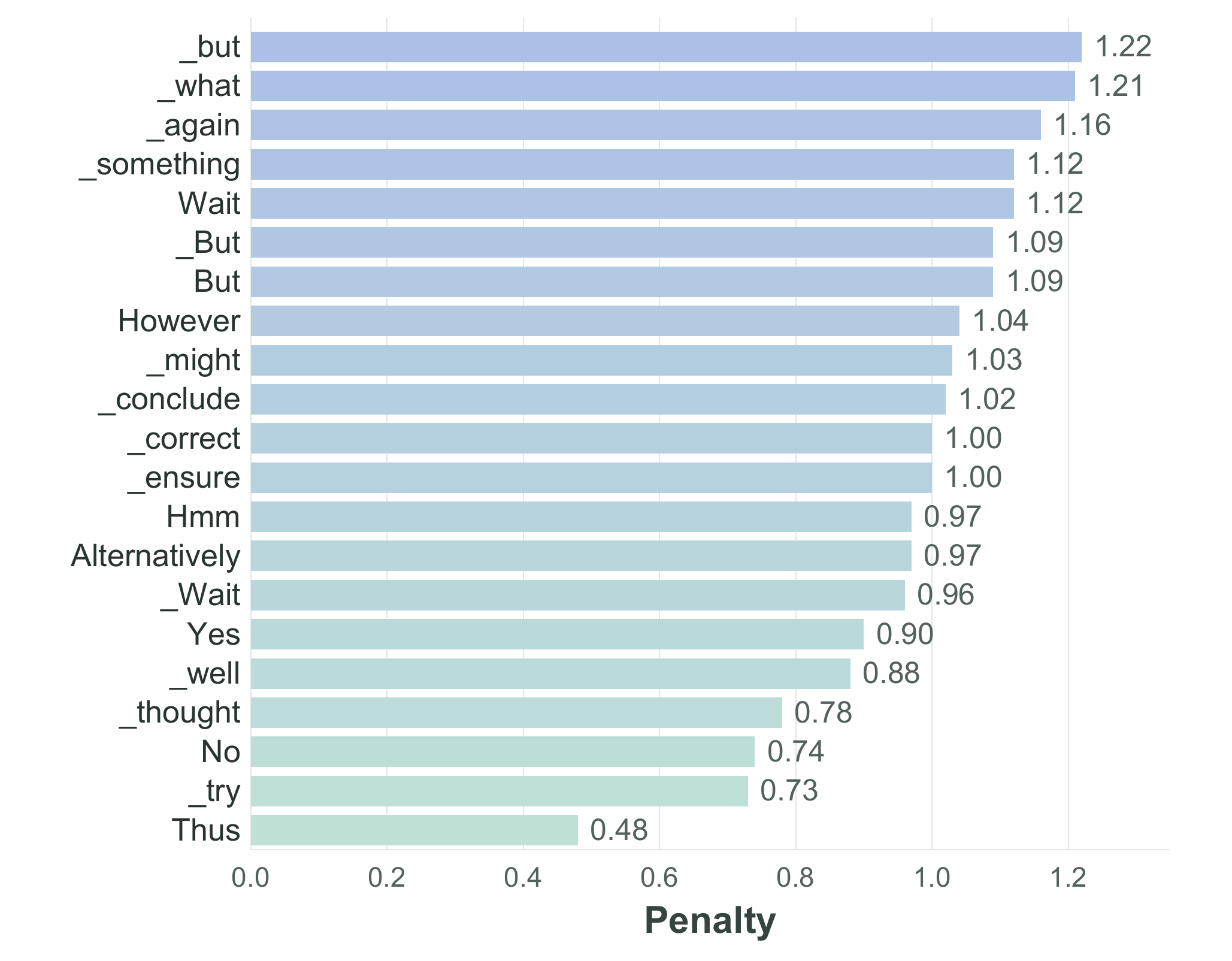}
        \par\smallskip
        {\small (b) Token-specific penalties\par}
    \end{minipage}
    \caption{Target tokens and calibrated penalties for Qwen-1.5B. (a) Word cloud of the 21 selected tokens, repeated for visualization. (b) Token-specific logit penalties in descending order.}
    \label{fig:entity-relation-statistics}
    \vspace{-5mm}
\end{figure}

To address these issues, recent studies have explored various approaches, which can be broadly categorized into optimization-based methods and inference-time interventions. Optimization-based approaches typically adopt sophisticated strategies, such as task-specific fine-tuning~\citep{li2025quantizationmeets}, full-precision model guidance~\citep{alimaskina2026extreme}, or dynamic precision adjustment~\citep{chen2026astro}, to recover the reasoning capability of quantized models. Recently, a lightweight decoding-time strategy~\citep{lotfi2026quantized} has been proposed to mitigate inefficient reasoning behaviors by penalizing manually selected overthinking markers. However, such marker-based strategies rely on predefined token lists and cannot adapt to model-specific quantization-induced reasoning changes, as overthinking tokens may vary across models. Moreover, tokens exhibit different degrees of reasoning deviation, making uniform penalties insufficient to capture token-level differences.

In this work, we propose Reasoning Analysis and Token-level Inference Optimization (\textbf{RATIO}), a framework that discovers model-specific overthinking tokens and performs token-specific calibration for quantized reasoning models. \textbf{RATIO} introduces Quantization-aware Reasoning Behavior Analysis (QRBA) to discover overthinking tokens through a two-stage process: Quantization-Sensitive Token Identification (QSTI), which identifies candidate tokens with token-level shifts, and Reasoning-context-aware Token Validation (RTV), which further selects tokens associated with inefficient reasoning behaviors based on actual quantized models' trajectories. Based on the selected tokens, Token-Specific Penalty Determination (TSPD) leverages the full-precision model as guidance to derive token-specific penalties according to token-level generation discrepancies. As shown in Fig.~\ref{fig:entity-relation-statistics}, QRBA identifies a set of model-specific reasoning tokens for Qwen-1.5B, while TSPD assigns a distinct penalty to each token. By adapting token selection and penalty strength to quantized models, \textbf{RATIO} suppresses unnecessary reasoning tokens while preserving or improving reasoning accuracy. As shown in Fig.~\ref{fig:fig1}, on DeepSeek-R1-Distill-Qwen-1.5B under GPTQ-W3, RATIO improves average accuracy by 3.58 percentage points and reduces CoT length by 25.7\%, outperforming the best fixed-penalty baseline on both metrics.

Our main contributions are summarized as follows:
\vspace{-1mm}
\begin{itemize}
\vspace{-1mm}
\item We propose \textbf{RATIO}, a training-free framework for quantized reasoning models that enables token-level calibration without additional inference overhead.
\vspace{-1mm}
\item We introduce Quantization-aware Reasoning Behavior Analysis (QRBA), which discovers model-specific overthinking tokens through token-level discrepancy analysis and reasoning-context-aware validation on actual reasoning trajectories.
\vspace{-1mm}
\item We develop Token-Specific Penalty Determination (TSPD), which leverages the full-precision model as a reference to derive token-specific penalties and calibrates quantized models' reasoning behaviors with tailored strengths.

\vspace{-1mm}
\item Extensive experiments across multiple reasoning benchmarks, model scales and quantization settings demonstrate that \textbf{RATIO} effectively reduces unnecessary reasoning while preserving or improving the reasoning accuracy, achieving a better accuracy-efficiency trade-off over the corresponding quantized baselines.

\end{itemize}

\section{Related Works}
\label{gen_inst}

\subsection{Post-Training Quantization}
Post-training quantization (PTQ) has become a widely adopted technique for efficient LLM deployment. By converting high-precision parameters into low-bit representations, PTQ reduces memory and computation costs without requiring additional training. Existing PTQ methods can be broadly categorized into mixed-precision, compensation-based, and transformation-based approaches. \textbf{Mixed-precision methods}~\citep{zhao2024atom,2025skim,huang_slim-llm_2025,saxena2025resq} adaptively allocate varying bit-widths based on quantization sensitivity. For example, Quik~\citep{QUIK} retains outlier channels at higher precision while quantizing most weights and activations to 4 bits. \textbf{Compensation-based methods}~\citep{li_gptaq_2025,kim_boa_nodate,arai_quantization_2026} such as GPTQ~\citep{frantar_gptq_2023} reduce reconstruction error by employing Hessian-based optimization to adjust full-precision weights. \textbf{Transformation-based methods}, including SmoothQuant~\citep{xiao_smoothquant_2024}, AWQ~\citep{lin_awq_2024}, and QuaRot~\citep{ashkboos_quarot_2024}, apply equivalent transformations or rotations to alleviate outliers and improve low-bit quantization robustness. Beyond methods developed for integer quantization, recent works~\citep{chen2025razer,bao2026soar,yan2026focus} have investigated microscaling floating-point formats such as MXFP4 and NVFP4. For instance, MR-GPTQ~\citep{mr-gptq2026} further improves FP4 quantization through discrete scaling search and block-wise Hadamard rotation. However, conventional quantization evaluation mainly focuses on accuracy and per-token generation speed, which is incomplete for reasoning models because the number of generated tokens is itself a major component of the total computational cost. 

\subsection{Effect of quantization on reasoning}
Recent studies have begun to investigate the effects of quantization on reasoning models. Empirical analyses~\citep{li2025quantizationmeets,liu2025quantizationhurts} report degradation in reasoning accuracy and show that its severity varies with the model, quantization setting, and task. Further work~\citep{lian2026quantizationinflates} finds that quantization can also increase the number of reasoning tokens, even when answer accuracy is preserved. To improve the reasoning capability of quantized models, existing approaches have explored different recovery and optimization strategies. Some methods~\citep{li2025quantizationmeets,zhang2026quantlrm} leverage additional training signals, such as task-specific adaptation or fine-tuning-guided quantization, to recover quantized reasoning performance. Other approaches focus on inference-time adaptation, where ASTRO~\citep{chen2026astro} reduces reasoning costs through adaptive precision allocation and termination strategies, while FP16 planning and loop rescue~\citep{alimaskina2026extreme} mitigate generation failures in extremely low-bit reasoning models. More closely related to our work, a recent decoding-time method~\citep{lotfi2026quantized} reduces unnecessary reasoning by penalizing a manually specified set of overthinking markers with a shared logit bias. However, such fixed marker-based strategies overlook how strongly a token is associated with inefficient reasoning in different contexts and how much its probability shifts under quantization across models. To address this, RATIO identifies model-specific reasoning tokens and assigns tailored penalties using full-precision guidance, enabling token-specific calibration without additional training.

\section{Method}
\subsection{Overview of RATIO}
\begin{figure*}[t]
  \centering
  \vspace{-1mm}
   \includegraphics[width=\textwidth]{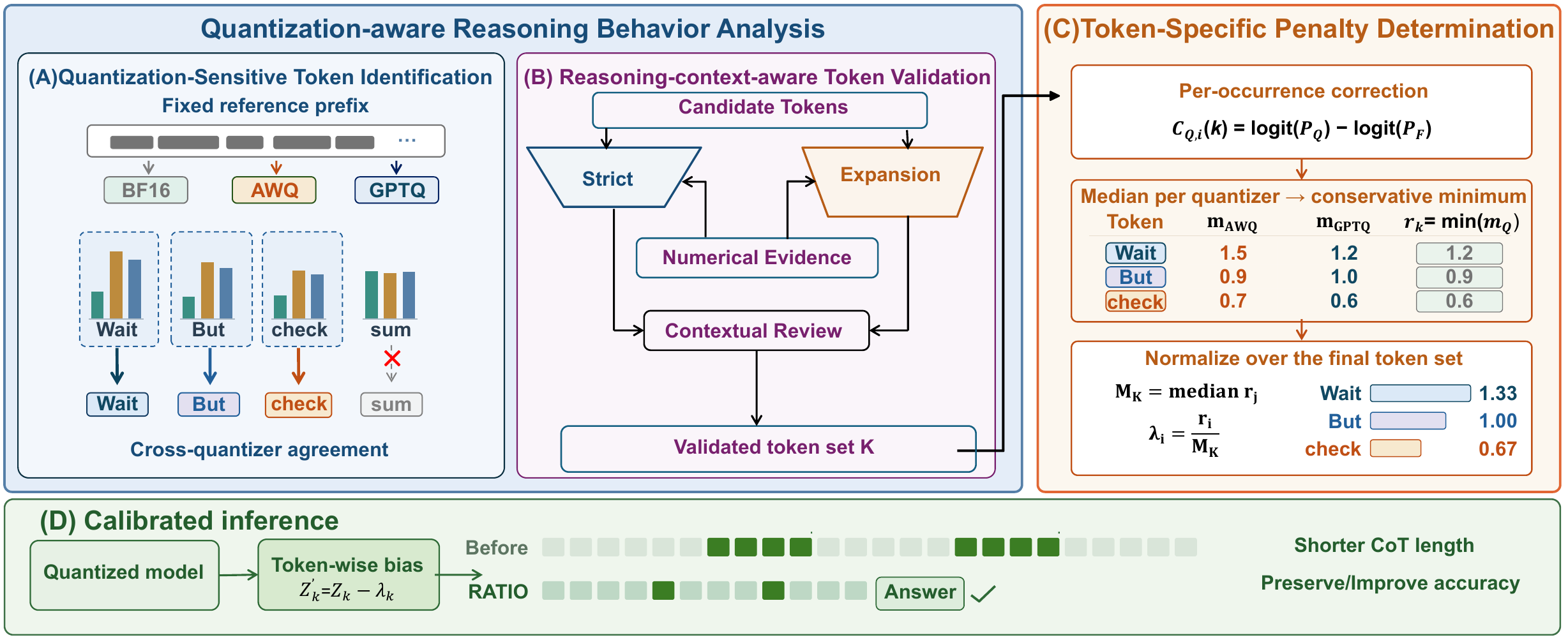}
   \vspace{-5mm}
    \caption{Overview of RATIO. The two panels on the left illustrate QRBA, which discovers model-specific quantization-sensitive reasoning tokens by combining token-level discrepancies with reasoning-context information. The rightmost panel presents TSPD, which leverages full-precision guidance to derive token-specific penalties for inference-time calibration.}
   \label{fig:overview}
   \vspace{-4mm}
\end{figure*}

In this section, we introduce our method RATIO, as illustrated in Fig.~\ref{fig:overview}. To analyze and mitigate quantization-induced changes in reasoning behavior, for each base model, RATIO compares its full-precision version with two quantized variants produced by AWQ and GPTQ. RATIO first performs Quantization-aware Reasoning Behavior Analysis (QRBA), as detailed in Sec.~\ref{sec:qrba}, to discover model-specific overthinking tokens by combining token-level discrepancies and reasoning-context information. Based on the identified tokens, RATIO further performs Token-Specific Penalty Determination (TSPD), as described in Sec.~\ref{sec:tspd}, to derive token-specific penalties for inference-time calibration under full-precision guidance.

\subsection{Quantization-aware Reasoning Behavior Analysis}
\label{sec:qrba}

\subsubsection{Quantization-Sensitive Token Identification.} To identify quantization-sensitive tokens, we perform a controlled comparison between full-precision and quantized models under identical reasoning contexts. Given a fixed reference reasoning trajectory, the full-precision model and its quantized counterparts are evaluated with the same input prefix using teacher forcing, eliminating the influence of different generation trajectories. For each reasoning position, we measure the next-token probability shift between a quantized model $Q$ and the full-precision model $F$:
\begin{equation}
\delta_{Q,i}(k)
=
\log p_Q(k|x_{<t_i})
-
\log p_F(k|x_{<t_i}),
\end{equation}
where $k$ denotes a token retained from the union of the two models' top-$p$ candidate sets at position $i$, and $x_{<t_i}$ represents their shared prefix before the $i$-th position. Further details of the candidate sets are provided in the Appendix~\ref{app:qsti_details}.

For each token $k$, we aggregate its occurrence-level probability shifts across reasoning positions and compute the average shift separately for each quantized model:
\begin{equation}
\bar{\delta}_{Q}(k)
=
\frac{1}{N_Q(k)}
\sum_i \delta_{Q,i}(k),
\end{equation}
where $N_Q(k)$ denotes the number of response positions where token $k$ is retained from the union of the top-\(p\) candidate sets for \(Q\) and \(F\). To reduce quantizer-specific effects, we compare the average shifts from AWQ and GPTQ and retain tokens with positive shifts in both models, subject to additional support and stability criteria detailed in the Appendix~\ref{app:qsti_details}, for subsequent reasoning-context-aware validation. Note that this stage only captures quantization-related token-level distribution changes; the identified shifts alone do not necessarily indicate inefficient reasoning behaviors.

\begin{figure}[t]
    \centering
    \includegraphics[width=\textwidth,trim=60bp 77bp 60bp 38bp,clip]{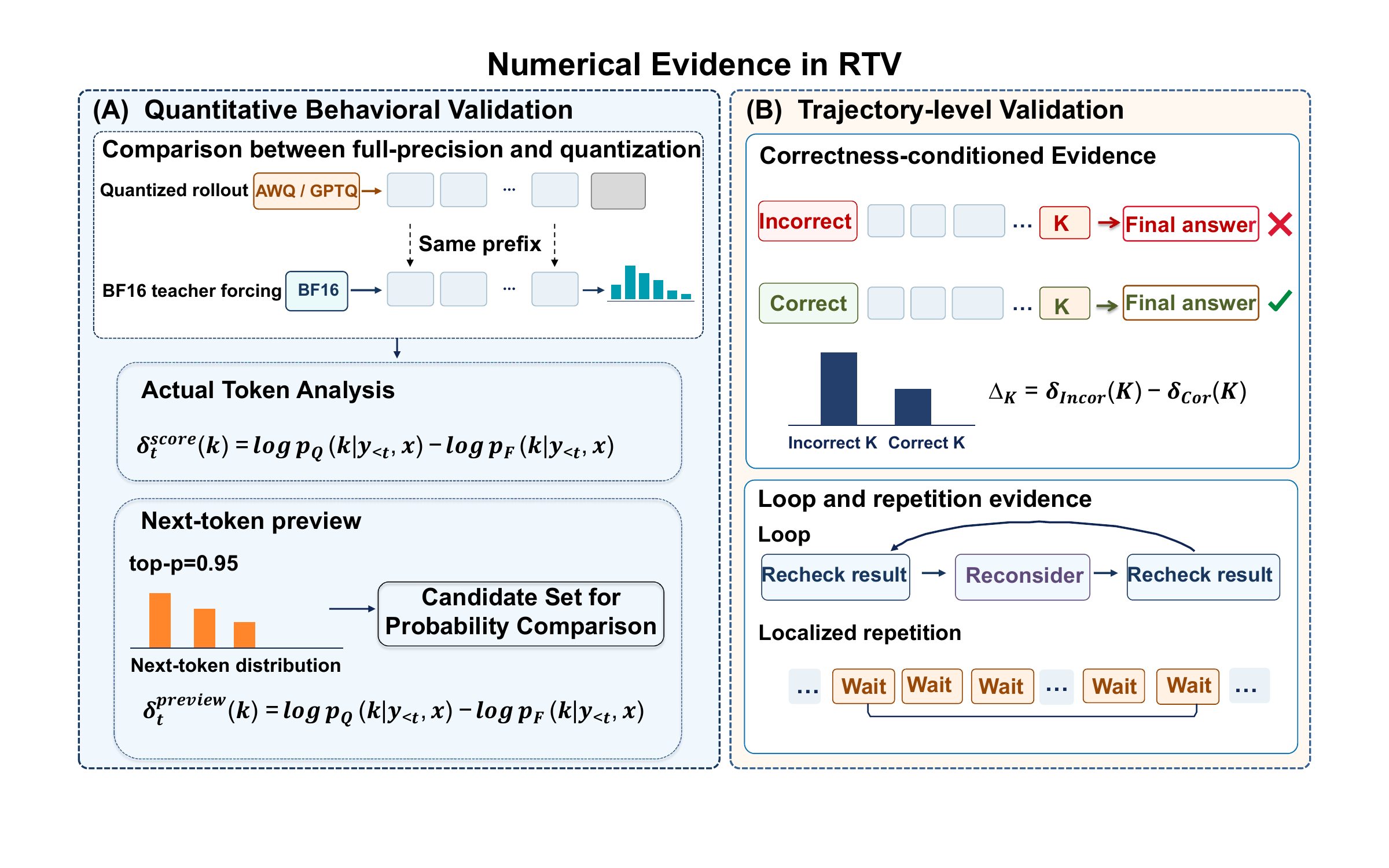}
    \vspace{-5mm}
    \caption{Numerical evidence construction in RTV. (A) Quantitative behavioral validation compares token probabilities under the same quantized prefixes. (B) Trajectory-level validation examines associations with incorrect answers and repetitive reasoning. Here, $Q$ and $F$ denote the quantized and full-precision models, respectively.}
    \label{fig:rtv-numerical-evidence}
    \vspace{-5mm}
\end{figure}
\subsubsection{Reasoning-context-aware Token Validation} Token shifts identified on fixed reference trajectories reveal how quantization changes next-token distributions, but they do not determine whether the affected tokens are associated with inefficient reasoning behaviors. We therefore further validate the tokens obtained from QSTI using freely generated reasoning trajectories from the quantized models.

For each quantized model, we first generate reasoning trajectories on quantized models, while an independent full-precision rollout on the same questions only provides correctness information. We then compare the quantized and full-precision models on the generated quantized trajectories: the quantized model follows its own generated trajectory, while the full-precision model evaluates the same prefixes through teacher forcing. This enables token-level probability comparison under identical reasoning states while avoiding the influence of different free-generation trajectories.

\textbf{Quantitative Behavioral Validation.} We first examine whether the quantization-related shifts identified on fixed trajectories persist during actual generation. For each token obtained from QSTI, we measure token-level probability deviations during quantized responses from two perspectives. Actual-token analysis evaluates the probability difference assigned to the token generated by the quantized model, while next-token preview analysis compares the top-\(p\) candidate distributions of the quantized and full-precision models at the same generation position. The former captures deviations of sampled tokens, while the latter captures tokens that are favored by quantization even when they are not selected during generation.

\textbf{Trajectory-level Validation.} Beyond quantitative behavioral validation, we further examine whether the identified tokens are associated with inefficient reasoning trajectories. Specifically, we analyze whether these tokens exhibit stronger associations with incorrect reasoning cases by comparing their occurrences between quantized-failure trajectories and successful reasoning trajectories. We additionally measure their association with repetitive reasoning patterns, including explicit loops and other forms of repetition. Together with quantitative behavioral validation, these trajectory-level statistics provide numerical evidence linking candidate tokens to inefficient reasoning outcomes. Figure~\ref{fig:rtv-numerical-evidence} illustrates the construction of these numerical signals. Detailed formulations of these trajectory-level measurements are provided in the Appendix~\ref{app:rtv_details}.

\textbf{Contextual Validation.} Both preceding components rely on numerical measurements of token shifts and trajectory outcomes. However, such evidence may still be insufficient to distinguish inefficient deliberation from useful reasoning operations, such as verification and self-correction. We therefore inspect the surrounding reasoning contexts of candidate tokens to better understand their roles within reasoning trajectories. Importantly, contextual validation is only applied after tokens satisfy the corresponding numerical criteria in each selection stage. It determines token retention or exclusion but does not introduce additional tokens beyond the numerically qualified candidates.

\textbf{Selection Strategy.} Based on the above validation signals, Reasoning-context-aware Token Validation (RTV) adopts a staged token selection strategy. Tokens that satisfy strong quantitative behavioral criteria and sufficient support across AWQ and GPTQ are first considered as high-confidence candidates, followed by contextual validation to determine their retention. For borderline tokens with weaker quantitative behavioral evidence, additional trajectory-level and contextual validation signals are considered to determine whether they should be retained. Through this process, RTV obtains a model-specific token set for subsequent Token-Specific Penalty Determination.

\subsection{Token-Specific Penalty Determination}
\label{sec:tspd}
After identifying the final quantization-sensitive reasoning token set, RATIO performs Token-Specific Penalty Determination (TSPD) to adjust the generation preference of each token during inference. Unlike previous approaches that apply a shared penalty, TSPD derives token-specific calibration penalties according to the deviation between quantized and full-precision models.

TSPD derives its calibration signal from the fixed reference trajectories used in QSTI. Let $K$ denote the final model-specific token set obtained by QRBA. Given a selected token $k\in K$, the full-precision model and each quantized model are evaluated under the same reference prefix $x_{<t_i}$ using teacher forcing. For a quantized model $Q$ and the full-precision model $F$, we obtain:
\begin{equation}
p_Q(k|x_{<t_i}), \qquad p_F(k|x_{<t_i}),
\end{equation}
where $Q$ denotes either the AWQ or GPTQ variant. Since TSPD only suppresses tokens whose preference is increased after quantization, we retain events satisfying:
\begin{equation}
p_Q(k|x_{<t_i})>p_F(k|x_{<t_i}).
\end{equation}
Instead of directly using probability differences, TSPD derives the correction magnitude in the logit space:
\begin{equation}
c_{Q,i}(k)
=
\operatorname{logit}(p_Q(k|x_{<t_i}))
-
\operatorname{logit}(p_F(k|x_{<t_i})),
\end{equation}
where
\begin{equation}
\operatorname{logit}(p)=\log\frac{p}{1-p}.
\end{equation}
The derivation of this logit-space correction is provided in the Appendix~\ref{app:tspd_derivation}. This quantity represents the token-level correction magnitude required to align the quantized preference of token $k$ with its full-precision preference.

We use the median of correction magnitudes across occurrences to obtain the token-level correction strength, and further combine the values from AWQ and GPTQ through conservative selection:
\begin{equation}
\begin{aligned}
m_Q(k) &= \operatorname{median}_{i:p_Q(k|x_{<t_i})>p_F(k|x_{<t_i})} c_{Q,i}(k),
&\quad
r_k &= \min\{m_{\mathrm{AWQ}}(k),m_{\mathrm{GPTQ}}(k)\}.
\end{aligned}
\end{equation}
Finally, we normalize the correction magnitudes within the selected token set:
\begin{equation}
M_K=\operatorname{median}_{j\in K}r_j,
\qquad
\lambda_k=\frac{r_k}{M_K}.
\end{equation}
During inference, the corresponding token logits are adjusted by:
\begin{equation}
z'_k=z_k-\lambda_k,\quad k\in K.
\end{equation}
By assigning different calibration strengths to different tokens, TSPD provides a lightweight inference-time mechanism to correct quantization-induced token preference shifts.

\begin{table}[t]
\centering
\caption{Accuracy (\%) and CoT length (k tokens) for BF16 and W3-quantized models using GPTQ or AWQ, with shared-penalty decoding (+~\cite{lotfi2026quantized}) or our method (+RATIO). The $\Delta$ columns report the accuracy change in percentage points and the relative CoT length change (\%) compared with the corresponding quantized model without intervention. For each model, the fixed-penalty baseline uses the penalty strength that achieves the highest mean accuracy across AWQ and GPTQ, averaged over the five benchmarks.}
\label{tab:main-results}
\resizebox{\textwidth}{!}{%
\setlength{\tabcolsep}{4.5pt}
\renewcommand{\arraystretch}{1.30}
\small
\begin{tabular}{cl ccccc c cc}
\toprule
 & & \textbf{AIME} & \textbf{MATH} & \textbf{GSM8K} & \textbf{GPQA} & \textbf{HumanEval} & \textbf{Avg.} & \textbf{$\Delta$Acc} & \textbf{$\Delta$Len\%} \\
\cmidrule(lr){9-9} \cmidrule(lr){10-10}
 & & \multicolumn{5}{c}{\footnotesize\textit{Acc (\%)\;/\;Len (k)}} & \footnotesize\textit{Acc\;/\;Len} & \multicolumn{2}{c}{\footnotesize\textit{vs. baseline}} \\
\midrule
 & \textbf{BF16} & \textbf{20.83}\,/\,\textbf{25.11} & \textbf{85.60}\,/\,\textbf{6.06} & \textbf{84.69}\,/\,\textbf{2.81} & \textbf{36.36}\,/\,\textbf{9.97} & \textbf{73.17}\,/\,\textbf{6.50} & \textbf{60.13}\,/\,\textbf{10.09} & -- & -- \\
\cline{2-10}
 & GPTQ & 4.17\,/\,51.53 & 54.00\,/\,24.27 & 68.76\,/\,9.54 & 31.82\,/\,23.67 & 10.37\,/\,39.31 & 33.82\,/\,29.66 & -- & -- \\
 & +~\cite{lotfi2026quantized} & 4.17\,/\,47.53 & 58.20\,/\,18.48 & 70.81\,/\,6.22 & 27.27\,/\,21.14 & 20.73\,/\,32.16 & 36.24\,/\,25.11 & $+2.42$ & $-15.34$ \\
\rowcolor{orange!10}
\cellcolor{white} & \textbf{+RATIO} & 5.00\,/\,42.55 & 61.40\,/\,14.06 & 71.49\,/\,3.11 & 30.81\,/\,20.64 & 18.29\,/\,29.90 & 37.40\,/\,22.05 & $+3.58$ & $-25.66$ \\
\cline{2-10}
 & AWQ & 7.50\,/\,55.53 & 44.80\,/\,36.13 & 61.18\,/\,22.16 & 23.23\,/\,36.56 & 17.68\,/\,39.81 & 30.88\,/\,38.04 & -- & -- \\
 & +~\cite{lotfi2026quantized} & 6.67\,/\,41.60 & 56.20\,/\,19.36 & 68.54\,/\,8.17 & 28.28\,/\,28.76 & 29.27\,/\,31.43 & 37.79\,/\,25.86 & $+6.91$ & $-32.02$ \\
\rowcolor{orange!10}
\cellcolor{white}\multirow{-7}{*}{\rotatebox{90}{Qwen-1.5B}} & \textbf{+RATIO} & 6.67\,/\,32.52 & 62.80\,/\,13.86 & 72.18\,/\,3.63 & 26.26\,/\,19.87 & 35.37\,/\,22.68 & 40.66\,/\,18.51 & $+9.78$ & $-51.34$ \\
\midrule
 & \textbf{BF16} & \textbf{42.50}\,/\,\textbf{14.27} & \textbf{93.60}\,/\,\textbf{3.83} & \textbf{91.96}\,/\,\textbf{1.50} & \textbf{52.02}\,/\,\textbf{8.60} & \textbf{85.98}\,/\,\textbf{4.87} & \textbf{73.21}\,/\,\textbf{6.61} & -- & -- \\
\cline{2-10}
 & GPTQ & 25.00\,/\,33.84 & 86.80\,/\,11.06 & 88.63\,/\,3.55 & 39.39\,/\,11.50 & 71.95\,/\,9.49 & 62.35\,/\,13.89 & -- & -- \\
 & +~\cite{lotfi2026quantized} & 24.17\,/\,34.57 & 84.80\,/\,9.25 & 88.25\,/\,2.43 & 42.93\,/\,11.74 & 81.10\,/\,6.70 & 64.25\,/\,12.94 & $+1.90$ & $-6.84$ \\
\rowcolor{orange!10}
\cellcolor{white} & \textbf{+RATIO} & 25.83\,/\,29.15 & 89.60\,/\,6.30 & 88.40\,/\,2.37 & 45.96\,/\,9.81 & 75.00\,/\,7.36 & 64.96\,/\,11.00 & $+2.61$ & $-20.81$ \\
\cline{2-10}
 & AWQ & 30.83\,/\,19.91 & 87.80\,/\,6.40 & 89.69\,/\,2.33 & 43.43\,/\,9.96 & 76.83\,/\,8.34 & 65.72\,/\,9.39 & -- & -- \\
 & +~\cite{lotfi2026quantized} & 31.67\,/\,17.62 & 89.00\,/\,5.50 & 90.14\,/\,1.85 & 46.46\,/\,9.16 & 81.10\,/\,7.76 & 67.67\,/\,8.38 & $+1.95$ & $-10.76$ \\
\rowcolor{orange!10}
\cellcolor{white}\multirow{-7}{*}{\rotatebox{90}{Qwen-7B}} & \textbf{+RATIO} & 32.50\,/\,16.41 & 88.60\,/\,5.07 & 89.46\,/\,1.82 & 45.96\,/\,8.45 & 81.10\,/\,4.98 & 67.52\,/\,7.34 & $+1.80$ & $-21.83$ \\
\midrule
 & \textbf{BF16} & \textbf{52.50}\,/\,\textbf{13.01} & \textbf{95.20}\,/\,\textbf{3.74} & \textbf{94.47}\,/\,\textbf{1.44} & \textbf{62.12}\,/\,\textbf{8.21} & \textbf{93.29}\,/\,\textbf{3.18} & \textbf{79.52}\,/\,\textbf{5.92} & -- & -- \\
\cline{2-10}
 & GPTQ & 36.67\,/\,14.92 & 92.20\,/\,4.06 & 92.34\,/\,1.42 & 48.48\,/\,7.18 & 93.90\,/\,4.23 & 72.72\,/\,6.36 & -- & -- \\
 & +~\cite{lotfi2026quantized} & 43.33\,/\,13.27 & 91.00\,/\,3.92 & 92.95\,/\,1.30 & 53.54\,/\,7.48 & 95.12\,/\,3.77 & 75.19\,/\,5.95 & $+2.47$ & $-6.45$ \\
\rowcolor{orange!10}
\cellcolor{white} & \textbf{+RATIO} & 35.00\,/\,11.55 & 91.80\,/\,3.21 & 92.95\,/\,1.10 & 57.58\,/\,6.19 & 92.68\,/\,2.74 & 74.00\,/\,4.96 & $+1.28$ & $-22.01$ \\
\cline{2-10}
 & AWQ & 39.17\,/\,14.29 & 93.20\,/\,4.52 & 93.33\,/\,1.77 & 54.55\,/\,8.65 & 92.68\,/\,4.50 & 74.59\,/\,6.75 & -- & -- \\
 & +~\cite{lotfi2026quantized} & 38.33\,/\,16.09 & 92.00\,/\,4.43 & 93.63\,/\,1.68 & 50.51\,/\,8.44 & 91.46\,/\,4.36 & 73.19\,/\,7.00 & $-1.40$ & $+3.70$ \\
\rowcolor{orange!10}
\cellcolor{white}\multirow{-7}{*}{\rotatebox{90}{Qwen-14B}} & \textbf{+RATIO} & 40.83\,/\,14.08 & 92.40\,/\,3.65 & 93.48\,/\,1.28 & 55.05\,/\,7.32 & 90.85\,/\,3.20 & 74.52\,/\,5.91 & $-0.07$ & $-12.44$ \\
\midrule
 & \textbf{BF16} & \textbf{35.83}\,/\,\textbf{16.26} & \textbf{89.00}\,/\,\textbf{4.75} & \textbf{87.87}\,/\,\textbf{1.81} & \textbf{46.97}\,/\,\textbf{9.20} & \textbf{81.10}\,/\,\textbf{3.88} & \textbf{68.15}\,/\,\textbf{7.18} & -- & -- \\
\cline{2-10}
 & GPTQ & 15.00\,/\,24.57 & 73.60\,/\,8.62 & 74.37\,/\,1.73 & 34.85\,/\,10.80 & 70.73\,/\,7.99 & 53.71\,/\,10.74 & -- & -- \\
 & +~\cite{lotfi2026quantized} & 15.00\,/\,24.81 & 73.20\,/\,6.30 & 75.06\,/\,1.38 & 37.88\,/\,10.60 & 68.90\,/\,7.11 & 54.01\,/\,10.04 & $+0.30$ & $-6.52$ \\
\rowcolor{orange!10}
\cellcolor{white} & \textbf{+RATIO} & 15.00\,/\,23.58 & 71.40\,/\,6.33 & 73.77\,/\,1.52 & 37.37\,/\,9.64 & 75.00\,/\,6.80 & 54.51\,/\,9.57 & $+0.80$ & $-10.89$ \\
\cline{2-10}
 & AWQ & 18.33\,/\,16.40 & 78.00\,/\,5.39 & 83.40\,/\,2.43 & 34.34\,/\,8.14 & 72.56\,/\,5.99 & 57.33\,/\,7.67 & -- & -- \\
 & +~\cite{lotfi2026quantized} & 15.83\,/\,12.93 & 75.40\,/\,5.78 & 82.11\,/\,1.93 & 33.33\,/\,7.82 & 73.78\,/\,4.94 & 56.09\,/\,6.68 & $-1.24$ & $-12.91$ \\
\rowcolor{orange!10}
\cellcolor{white}\multirow{-7}{*}{\rotatebox{90}{Llama-8B}} & \textbf{+RATIO} & 16.67\,/\,17.87 & 76.80\,/\,4.95 & 83.32\,/\,2.06 & 36.36\,/\,8.03 & 73.78\,/\,3.77 & 57.39\,/\,7.33 & $+0.06$ & $-4.43$ \\
\midrule
 & \textbf{BF16} & \textbf{71.67}\,/\,\textbf{16.91} & \textbf{97.40}\,/\,\textbf{5.35} & \textbf{94.39}\,/\,\textbf{2.32} & \textbf{52.02}\,/\,\textbf{6.60} & \textbf{90.85}\,/\,\textbf{4.60} & \textbf{81.27}\,/\,\textbf{7.15} & -- & -- \\
\cline{2-10}
 & GPTQ & 16.67\,/\,32.10 & 81.20\,/\,12.05 & 90.52\,/\,3.41 & 35.35\,/\,15.30 & 54.88\,/\,20.20 & 55.72\,/\,16.61 & -- & -- \\
 & +~\cite{lotfi2026quantized} & 20.00\,/\,31.08 & 80.20\,/\,11.70 & 90.90\,/\,3.27 & 32.32\,/\,15.19 & 57.93\,/\,18.36 & 56.27\,/\,15.92 & $+0.55$ & $-4.15$ \\
\rowcolor{orange!10}
\cellcolor{white} & \textbf{+RATIO} & 19.17\,/\,28.53 & 80.40\,/\,10.72 & 90.52\,/\,2.98 & 35.86\,/\,12.80 & 70.12\,/\,13.65 & 59.21\,/\,13.74 & $+3.49$ & $-17.28$ \\
\cline{2-10}
 & AWQ & 26.67\,/\,23.90 & 85.60\,/\,7.94 & 89.69\,/\,3.65 & 38.38\,/\,9.62 & 71.95\,/\,12.28 & 62.46\,/\,11.48 & -- & -- \\
 & +~\cite{lotfi2026quantized} & 25.00\,/\,22.75 & 87.40\,/\,7.30 & 91.43\,/\,3.11 & 40.91\,/\,10.34 & 76.22\,/\,10.82 & 64.19\,/\,10.87 & $+1.73$ & $-5.31$ \\
\rowcolor{orange!10}
\cellcolor{white}\multirow{-7}{*}{\rotatebox{90}{Qwen3-4B}} & \textbf{+RATIO} & 26.67\,/\,22.36 & 86.60\,/\,7.01 & 90.60\,/\,2.95 & 37.88\,/\,8.81 & 76.22\,/\,8.89 & 63.59\,/\,10.00 & $+1.13$ & $-12.89$ \\
\bottomrule
\end{tabular}%
}
\end{table}

\section{Experiments}
\subsection{Experimental Setup}
\label{experiment}
\textbf{Models and Quantization Settings.} We evaluate RATIO on five reasoning models, including DeepSeek-R1-Distill-Qwen-1.5B, DeepSeek-R1-Distill-Qwen-7B, DeepSeek-R1-Distill-Qwen-14B, DeepSeek-R1-Distill-Llama-8B, and Qwen3-4B~\citep{qwen3technicalreport} in thinking mode. The three Qwen-based distilled models and DeepSeek-R1-Distill-Llama-8B are built on the Qwen2.5~\citep{yang2024qwen25} and Llama 3.1~\citep{grattafiori_llama_2024}
architectures. For each model, we consider the full-precision BF16 model and two representative post-training quantization settings: AWQ~\citep{lin_awq_2024} and GPTQ~\citep{frantar_gptq_2023}. Both AWQ and GPTQ use 3-bit weight quantization with a group size of 128. Additional details about quantization configurations are provided in the Appendix~\ref{app:implementation}.

\textbf{Dataset Analyses.} We collect reasoning trajectories from several reasoning datasets for token identification and validation. For QSTI, we use fixed MATH-CoT reference trajectories~\citep{hendrycks2021math} to provide controlled reasoning contexts. For RTV, we generate reasoning trajectories on AIME~\citep{dekoninck2026matharena}, GPQA-Diamond~\citep{rein2024gpqa}, MATH-500~\citep{math500}, and GSM8K~\citep{cobbe2021training}, with 50 questions randomly selected from each benchmark for each quantized model. The analysis subsets of MATH-500 and GSM8K are sampled from their corresponding training splits. During RTV, the models are provided only with the input questions, without access to any reference answers.

\textbf{Evaluation and baseline.} We evaluate RATIO on five reasoning benchmarks, including AIME~\citep{dekoninck2026matharena}, GPQA-Diamond~\citep{rein2024gpqa}, MATH-500~\citep{math500}, GSM8K~\citep{cobbe2021training}, and HumanEval~\citep{chen2021evaluating}. We report answer accuracy to measure reasoning performance. To evaluate inference efficiency, we additionally measure the generated chain-of-thought length, defined as the number of generated tokens before the final answer. We compare RATIO with quantized baselines without calibration and the decoding-time intervention method~\citep{lotfi2026quantized} based on manually specified overthinking markers. For fixed-penalty baselines, the same penalty strength is applied to both AWQ and GPTQ for each model, consistent with the shared calibration setting of RATIO. All experiments use default decoding with $T=0.6$ and top-$p=0.95$, following the setting used by prior work~\citep{liu2025quantizationhurts} on quantization of reasoning models. All experiments are conducted on NVIDIA RTX A6000 GPUs, each with 48GB of memory, and use a maximum generation budget of 65,536 tokens.
\par
\vspace*{2mm}
\subsection{Main Results}
\textbf{Overall Accuracy.} We first evaluate the impact of RATIO on reasoning accuracy. As shown in Table~\ref{tab:main-results}, low-bit quantization leads to accuracy degradation, while the effectiveness of the fixed-penalty method varies across models and quantizers. In contrast, RATIO generally preserves or improves the accuracy of quantized models. For example, on Qwen-1.5B, RATIO improves the average accuracy of AWQ-W3 from 30.88\% to 40.66\% and that of GPTQ-W3 from 33.82\% to 37.40\%. Similar improvements are observed on larger models under different quantization settings. On Qwen3-4B under GPTQ-W3, RATIO improves average accuracy from 55.72\% to 59.21\%. These results demonstrate that RATIO effectively mitigates quantization-induced reasoning degradation across diverse reasoning models.

\textbf{Reasoning Efficiency.} We further evaluate the effect of RATIO on reasoning length. As shown in Table~\ref{tab:main-results}, RATIO consistently reduces the average CoT length of the evaluated models under both AWQ and GPTQ. On Qwen-1.5B, for example, it achieves reductions of 51.34\% under AWQ-W3 and 25.66\% under GPTQ-W3 relative to their uncalibrated counterparts. Reductions are also observed across the other model scales, demonstrating the general effectiveness of RATIO in controlling reasoning length. Moreover, RATIO generally produces shorter reasoning trajectories than the fixed-penalty baseline, improving the inference efficiency of quantized reasoning models. For instance, on Qwen-7B under GPTQ-W3, RATIO reduces average CoT length by 14.99\% relative to the fixed-penalty baseline while also achieving higher average accuracy. These results support RATIO's effectiveness in reducing CoT length across models and quantizers.

\begin{figure}[t]
    \centering
    \begin{minipage}[t]{0.495\textwidth}
        \centering
        \includegraphics[width=\linewidth,height=1.75in,keepaspectratio,trim=22bp 12bp 8bp 8bp,clip]{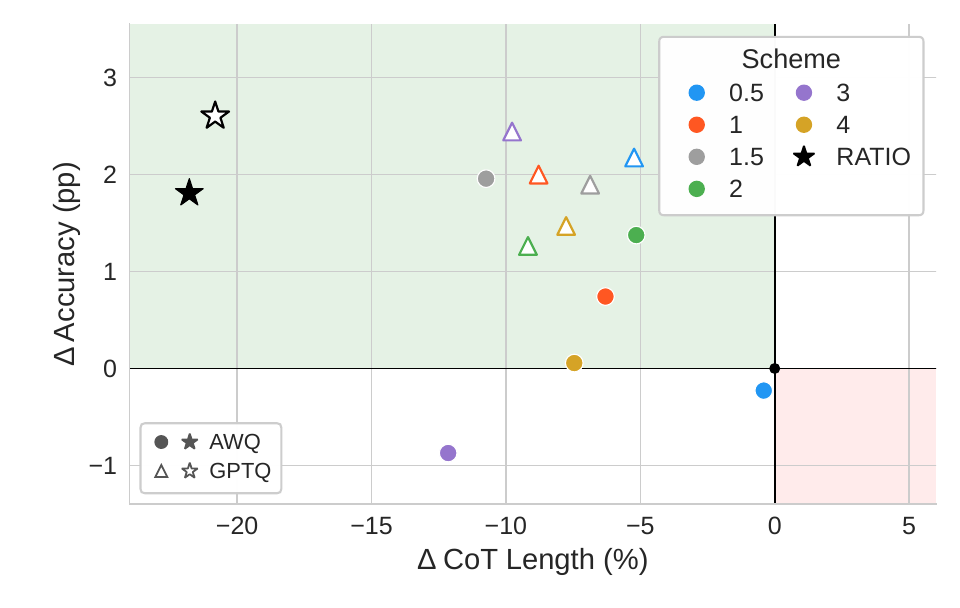}
        \par\smallskip
        {\small (a) Accuracy--efficiency trade-off\par}
    \end{minipage}\hfill
    \begin{minipage}[t]{0.495\textwidth}
        \centering
        \includegraphics[width=\linewidth,height=1.75in,keepaspectratio,trim=12bp 22bp 7bp 11bp,clip]{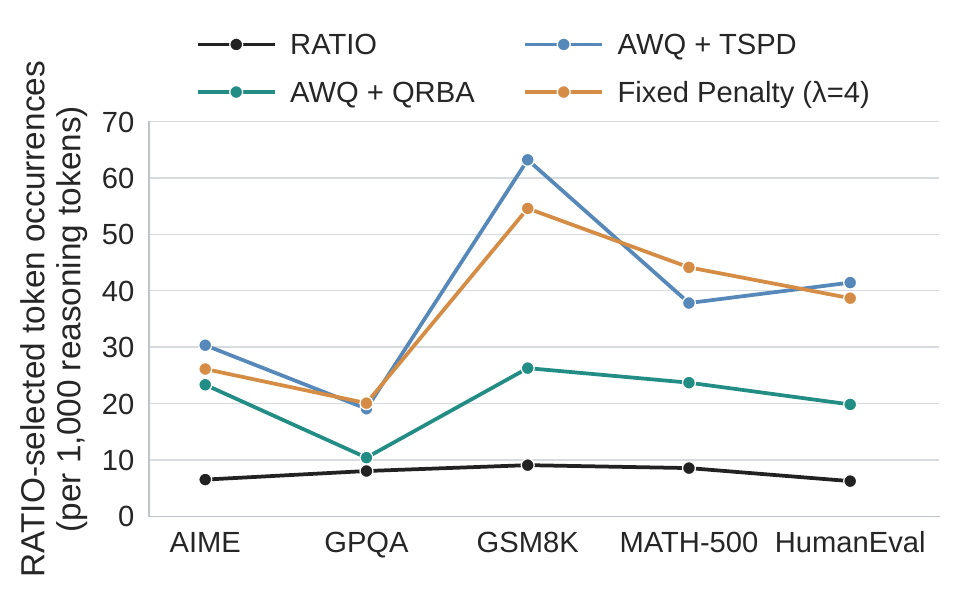}
        \par\smallskip
        {\small (b) Selected-token occurrence rates\par}
    \end{minipage}
    \caption{Accuracy–efficiency trade-off and selected-token occurrence analysis. (a) Changes in average accuracy and CoT length for fixed-penalty decoding with different penalty strengths and RATIO on Qwen-7B, measured relative to the corresponding uncalibrated AWQ-W3 and GPTQ-W3 baselines. RATIO occupies the upper-left region (shorter CoT and higher accuracy) under both quantizers. (b) Benchmark-wise occurrence rates of QRBA-selected tokens on Qwen-1.5B with AWQ-W3 under RATIO, its component ablations, and the fixed-penalty baseline. RATIO consistently produces lower occurrence rates across benchmarks, showing that it effectively suppresses the selected target tokens during reasoning.}
    \label{fig:accuracy-efficiency-tradeoff}
    \vspace{-4mm}
\end{figure}
Overall, RATIO achieves a more favorable accuracy–efficiency trade-off for quantized reasoning models. As illustrated in Fig.~\ref{fig:accuracy-efficiency-tradeoff}, RATIO substantially shortens reasoning trajectories while generally preserving or improving accuracy over uncalibrated quantized models, providing a better overall balance than fixed-penalty strategies. These results demonstrate the effectiveness of RATIO for accurate and efficient quantized reasoning.

\begin{table}[!htbp]
\centering
\caption{Effect of QRBA and TSPD. Component-wise ablation on Qwen-1.5B under AWQ-W3 and GPTQ-W3. QRBA applies uniform penalties to the identified tokens, while TSPD applies token-specific penalties to the manually selected markers from prior work. Each entry reports accuracy (\%) and CoT length (k tokens), with $\Delta$ columns showing changes relative to the corresponding quantized baseline.}
\label{tab:ablation}
\resizebox{\textwidth}{!}{%
\setlength{\tabcolsep}{4.5pt}
\renewcommand{\arraystretch}{1.25}
\small
\begin{tabular}{cl ccccc c cc}
\toprule
 & & \textbf{GPQA} & \textbf{AIME} & \textbf{GSM8K} & \textbf{MATH} & \textbf{HumanEval} & \textbf{Avg.} & \textbf{$\Delta$Acc} & \textbf{$\Delta$Len\%} \\
\cmidrule(lr){9-9} \cmidrule(lr){10-10}
 & & \multicolumn{5}{c}{\footnotesize\textit{Acc (\%)\;/\;Len (k)}} & \footnotesize\textit{Acc\;/\;Len} & \multicolumn{2}{c}{\footnotesize\textit{vs. baseline}} \\
\midrule
 & AWQ & 23.23\,/\,36.56 & 7.50\,/\,55.53 & 61.18\,/\,22.16 & 44.80\,/\,36.13 & 17.68\,/\,39.81 & 30.88\,/\,38.04 & -- & -- \\
 & QRBA ($\lambda=0.5$) & 29.80\,/\,32.37 & 4.17\,/\,48.73 & 69.29\,/\,9.17 & 59.40\,/\,20.21 & 34.76\,/\,29.86 & 39.48\,/\,28.07 & $+8.60$ & $-26.21$ \\
 & QRBA ($\lambda=1.0$) & 29.29\,/\,26.25 & 4.17\,/\,42.11 & 70.36\,/\,4.08 & 62.60\,/\,14.10 & 31.10\,/\,26.39 & 39.50\,/\,22.59 & $+8.62$ & $-40.62$ \\
 & TSPD & 20.71\,/\,31.74 & 7.50\,/\,47.74 & 66.19\,/\,10.20 & 59.00\,/\,20.21 & 33.54\,/\,28.25 & 37.39\,/\,27.63 & $+6.51$ & $-27.37$ \\
\rowcolor{orange!10}
\cellcolor{white} & \textbf{RATIO} & 26.26\,/\,19.87 & 6.67\,/\,32.52 & 72.18\,/\,3.63 & 62.80\,/\,13.86 & 35.37\,/\,22.68 & 40.66\,/\,18.51 & $+9.78$ & $-51.34$ \\
\cline{2-10}
 & GPTQ & 31.82\,/\,23.67 & 4.17\,/\,51.53 & 68.76\,/\,9.54 & 54.00\,/\,24.27 & 10.37\,/\,39.31 & 33.82\,/\,29.66 & -- & -- \\
 &  QRBA ($\lambda=0.5$) & 30.30\,/\,24.70 & 5.83\,/\,48.91 & 72.18\,/\,4.79 & 57.20\,/\,18.74 & 17.68\,/\,31.72 & 36.64\,/\,25.77 & $+2.82$ & $-13.12$ \\
 & QRBA ($\lambda=1.0$) & 30.30\,/\,22.81 & 6.67\,/\,40.04 & 72.25\,/\,3.13 & 61.00\,/\,15.29 & 23.17\,/\,28.29 & 38.68\,/\,21.91 & $+4.86$ & $-26.13$ \\
 & TSPD & 33.84\,/\,19.76 & 2.50\,/\,46.07 & 71.04\,/\,5.93 & 57.80\,/\,19.63 & 19.51\,/\,32.20 & 36.94\,/\,24.72 & $+3.12$ & $-16.66$ \\
\rowcolor{orange!10}
\cellcolor{white}\multirow{-10}{*}{\rotatebox{90}{Qwen-1.5B}} & \textbf{RATIO} & 30.81\,/\,20.64 & 5.00\,/\,42.55 & 71.49\,/\,3.11 & 61.40\,/\,14.06 & 18.29\,/\,29.90 & 37.40\,/\,22.05 & $+3.58$ & $-25.66$ \\
\bottomrule
\end{tabular}%
}
\vspace{-5mm}
\end{table}

\vspace{-2mm}
\subsection{Ablation Study}
\vspace{-1mm}
\textbf{Effect of QRBA.} We first assess the contribution of QRBA by replacing the model-specific token set identified by QRBA with the 50 manually selected overthinking markers from prior work~\citep{lotfi2026quantized}, while retaining TSPD. As shown in Table~\ref{tab:ablation}, this variant consistently underperforms the complete RATIO framework. For example, under AWQ-W3, this variant reduces average accuracy by 3.27 percentage points and increases average CoT length by 49.27\% compared with the complete RATIO framework. These results demonstrate the importance of QRBA in identifying model-specific tokens that can be effectively calibrated.

\textbf{Effect of TSPD.} We then replace TSPD with a uniform penalty applied to all tokens identified by QRBA. Overall, the uniform-penalty variants yield a less favorable accuracy--efficiency trade-off than the complete RATIO framework, and their performance varies between AWQ and GPTQ. For example, under AWQ-W3, the strongest uniform-penalty variant reduces average accuracy by 1.16 percentage points and increases average CoT length by 22.04\% compared with the complete RATIO framework. Also, a uniform penalty of $\lambda=1$ yields lower accuracy and longer CoT than RATIO under AWQ-W3, but higher accuracy and slightly shorter CoT under GPTQ-W3. These results show that token-specific penalties provide more reliable calibration than applying a shared penalty strength to all selected tokens.

Overall, these results highlight the complementary roles of model-specific token selection and token-specific penalty determination. As shown in Fig.~\ref{fig:accuracy-efficiency-tradeoff}(b), RATIO produces lower occurrence rates of the selected tokens than either component ablation and the fixed-penalty baseline across all five benchmarks, further illustrating their combined effect on reasoning behavior.
\vspace{-4mm}
\section{Discussion and Future works}
\vspace{-4mm}
While RATIO assigns token-specific penalties based on model-level quantization behaviors, these penalties remain static during inference. However, the same token may play different roles under different reasoning states, and a fixed penalty may not always be optimal for every occurrence. A promising future direction is to develop dynamic token calibration strategies that adapt the penalty strength according to the current reasoning context. For example, penalties can be relaxed during early occurrences of a token to preserve useful reasoning behaviors, while being strengthened when the token appears repeatedly within a short context window to suppress potential repetitive reasoning patterns. This would allow calibration to distinguish useful self-correction from repeated reconsideration that contributes little to solving the problem. Such state-aware calibration may further improve the balance between reasoning efficiency and model capability.
\vspace{-4mm}
\section{Conclusion}
\vspace{-4mm}
In this paper, we investigate the challenges of applying post-training quantization to reasoning models, and focus on the limitations of fixed token-level interventions. To address these issues, we propose RATIO, a framework that identifies model-specific target tokens and assigns each a tailored penalty. RATIO introduces Quantization-aware Reasoning Behavior Analysis (QRBA) to identify quantization-sensitive reasoning tokens through token-level distribution analysis and reasoning-context validation, and further develops Token-Specific Penalty Determination (TSPD) to derive token-specific penalties with full-precision guidance. Extensive experiments across multiple reasoning benchmarks and quantization settings demonstrate that RATIO effectively reduces reasoning length while preserving or improving accuracy, achieving a favorable accuracy-efficiency trade-off compared with existing token-level intervention strategies. This work establishes a behavior-aware calibration framework for quantized reasoning models, advancing efficient low-bit reasoning and providing a foundation for future research.

\bibliography{iclr2027_conference}
\bibliographystyle{iclr2027_conference}

\clearpage
\appendix
\section{More Experimental Results}
\label{app:experimental_results}

\subsection{Performance on Other Quantization Methods}
To evaluate whether RATIO remains effective beyond weight-only quantization, we further apply it to DeepSeek-R1-Distill-Qwen-1.5B quantized with FlatQuant~\citep{sun2024flatquant} under the W4A4KV4 setting. As shown in Table~\ref{tab:flatquant-results}, RATIO improves average accuracy from 45.27\% to 48.48\% while reducing average CoT length by 12.90\%. It also provides a more favorable accuracy--efficiency trade-off than fixed-penalty decoding, demonstrating its applicability to the joint low-bit quantization of weights, activations, and the KV cache.

\begin{table}[H]
\centering
\caption{Performance of RATIO on Qwen-1.5B with FlatQuant W4A4KV4. Accuracy (\%) and CoT length (k tokens) are reported as Acc/Len. The $\Delta$ columns report changes relative to uncalibrated FlatQuant. For each metric, the highest accuracy and shortest CoT length are in bold.}
\label{tab:flatquant-results}
\resizebox{\textwidth}{!}{%
\setlength{\tabcolsep}{4.5pt}
\renewcommand{\arraystretch}{1.25}
\small
\begin{tabular}{cl ccccc c cc}
\toprule
 & & \textbf{GPQA} & \textbf{AIME} & \textbf{GSM8K} & \textbf{MATH} & \textbf{HumanEval} & \textbf{Avg.} & \textbf{$\Delta$Acc} & \textbf{$\Delta$Len\%} \\
\cmidrule(lr){9-9} \cmidrule(lr){10-10}
 & & \multicolumn{5}{c}{\footnotesize\textit{Acc (\%)\;/\;Len (k)}} & \footnotesize\textit{Acc\;/\;Len} & \multicolumn{2}{c}{\footnotesize\textit{vs. baseline}} \\
\midrule
 & FlatQuant & \textbf{34.34}\,/\,25.68 & 8.33\,/\,\textbf{37.20} & 74.37\,/\,4.06 & 65.40\,/\,16.47 & 43.90\,/\,14.64 & 45.27\,/\,19.61 & -- & -- \\
 & +~\cite{lotfi2026quantized} & 30.30\,/\,\textbf{20.81} & \textbf{11.67}\,/\,39.26 & \textbf{75.66}\,/\,2.83 & 69.80\,/\,13.30 & 48.17\,/\,\textbf{11.75} & 47.12\,/\,17.59 & $+1.85$ & $-10.30$ \\
\rowcolor{orange!10}
\cellcolor{white}\multirow{-3}{*}{\rotatebox{90}{Qwen-1.5B}} & \textbf{+RATIO} & 30.81\,/\,22.16 & \textbf{11.67}\,/\,37.77 & 75.51\,/\,\textbf{1.69} & \textbf{73.20}\,/\,\textbf{11.87} & \textbf{51.22}\,/\,11.90 & \textbf{48.48}\,/\,\textbf{17.08} & $\mathbf{+3.21}$ & $\mathbf{-12.90}$ \\
\bottomrule
\end{tabular}%
}
\end{table}

\section{Additional Experimental Details}
\label{app:experimental_details}

\subsection{Selected Tokens and Penalties}
Tables~\ref{tab:selected-tokens-penalties} and~\ref{tab:qwen3-selected-tokens-penalties} report the final model-specific token sets produced by QRBA and their normalized TSPD penalties. The same model-specific token set and penalty values are applied to its AWQ-W3 and GPTQ-W3 variants.

\begin{table}[H]
\centering
\caption{Model-specific tokens and normalized penalties used by RATIO. A leading underscore denotes a whitespace prefix in the decoded token.}
\label{tab:selected-tokens-penalties}
\setlength{\tabcolsep}{4.5pt}
\renewcommand{\arraystretch}{1.08}
\small
\begin{tabular}{@{}lc lc lc lc@{}}
\toprule
\multicolumn{2}{c}{\textbf{Qwen-1.5B}} &
\multicolumn{2}{c}{\textbf{Qwen-7B}} &
\multicolumn{2}{c}{\textbf{Llama-8B}} &
\multicolumn{2}{c}{\textbf{Qwen-14B}} \\
\cmidrule(lr){1-2}\cmidrule(lr){3-4}\cmidrule(lr){5-6}\cmidrule(lr){7-8}
\textbf{Token} & $\boldsymbol{\lambda_k}$ &
\textbf{Token} & $\boldsymbol{\lambda_k}$ &
\textbf{Token} & $\boldsymbol{\lambda_k}$ &
\textbf{Token} & $\boldsymbol{\lambda_k}$ \\
\midrule
\texttt{\_but}           & 1.22 & \texttt{\_but}       & 1.06 & \texttt{\_back}      & 1.00 & \texttt{\_but}       & 1.00 \\
\texttt{\_what}          & 1.21 & \texttt{\_back}      & 0.58 & \texttt{\_try}       & 0.90 & \texttt{\_what}      & 1.04 \\
\texttt{\_try}           & 0.73 & \texttt{\_check}     & 0.48 & \texttt{\_well}      & 0.86 & \texttt{\_think}     & 0.71 \\
\texttt{\_again}         & 1.16 & \texttt{\_But}       & 1.15 & \texttt{\_check}     & 0.93 & \texttt{\_But}       & 1.23 \\
\texttt{\_well}          & 0.88 & \texttt{\_something} & 1.09 & \texttt{\_But}       & 1.12 & \texttt{\_something} & 1.57 \\
\texttt{\_But}           & 1.09 & \texttt{\_might}     & 0.75 & \texttt{\_something} & 1.34 & \texttt{\_question}  & 0.80 \\
\texttt{\_something}     & 1.12 & \texttt{\_question}  & 1.32 & \texttt{\_thought}   & 0.85 & \texttt{\_actually}  & 1.05 \\
\texttt{\_might}         & 1.03 & \texttt{\_wait}      & 0.92 & \texttt{\_question}  & 1.10 & \texttt{\_wait}      & 0.92 \\
\texttt{No}               & 0.74 & \texttt{But}          & 1.00 & \texttt{\_actually}  & 0.95 & \texttt{But}          & 1.14 \\
\texttt{\_thought}       & 0.78 & \texttt{\_trying}    & 0.64 & \texttt{\_wait}      & 0.96 & \texttt{\_correct}   & 0.92 \\
\texttt{But}              & 1.09 & \texttt{\_ensure}    & 0.67 & \texttt{But}          & 1.07 & \texttt{\_trying}    & 0.71 \\
\texttt{\_correct}       & 1.00 & \texttt{but}          & 1.04 & \texttt{\_correct}   & 1.13 & \texttt{\_seems}     & 0.88 \\
\texttt{\_ensure}        & 1.00 & \texttt{\_Wait}      & 1.18 & \texttt{\_seems}     & 1.11 & \texttt{\_maybe}     & 0.77 \\
\texttt{Yes}              & 0.90 & \texttt{Wait}         & 1.12 & \texttt{\_maybe}     & 0.93 & \texttt{\_confirm}   & 0.69 \\
\texttt{However}          & 1.04 & \texttt{\_Thus}      & 0.37 & \texttt{\_perhaps}   & 0.96 & \texttt{\_perhaps}   & 0.96 \\
\texttt{\_Wait}          & 0.96 & \texttt{Thus}         & 0.36 & \texttt{\_Wait}      & 1.14 & \texttt{\_Wait}      & 1.33 \\
\texttt{Wait}             & 1.12 & \texttt{Hmm}          & 1.03 & \texttt{something}    & 1.29 & \texttt{Wait}         & 1.32 \\
\texttt{\_conclude}      & 1.02 &                         &      &                       &      & \texttt{something}    & 1.78 \\
\texttt{Thus}             & 0.48 &                         &      &                       &      & \texttt{Thus}         & 0.36 \\
\texttt{Hmm}              & 0.97 &                         &      &                       &      & \texttt{Hmm}          & 1.23 \\
\texttt{Alternatively}    & 0.97 &                         &      &                       &      & \texttt{\_Hmm}       & 1.25 \\
\bottomrule
\end{tabular}
\end{table}

\begin{table}[H]
\centering
\caption{Selected tokens and normalized penalties for Qwen3-4B in thinking mode. A leading underscore denotes a whitespace prefix in the decoded token.}
\label{tab:qwen3-selected-tokens-penalties}
\setlength{\tabcolsep}{4.5pt}
\renewcommand{\arraystretch}{1.08}
\small
\begin{tabular}{@{}lc@{\hspace{3em}}lc@{}}
\toprule
\textbf{Token} & $\boldsymbol{\lambda_k}$ & \textbf{Token} & $\boldsymbol{\lambda_k}$ \\
\midrule
\texttt{\_but} & 1.11 & \texttt{\_question} & 1.11 \\
\texttt{\_what} & 1.03 & \texttt{But} & 1.06 \\
\texttt{\_back} & 0.83 & \texttt{\_However} & 0.94 \\
\texttt{\_think} & 0.81 & \texttt{but} & 1.00 \\
\texttt{\_But} & 1.05 & \texttt{\_perhaps} & 0.84 \\
\texttt{\_something} & 1.00 & \texttt{Wait} & 1.02 \\
\texttt{\_looking} & 0.82 & \texttt{\_mistake} & 0.79 \\
\bottomrule
\end{tabular}
\end{table}

\subsection{Quantization and Implementation Details}
\label{app:implementation}
\textbf{Quantization Settings.} We provide the detailed quantization and inference configurations used in our experiments. For AWQ, we follow the standard AWQ recipe and use 128 sequences with a sequence length of 512 sampled from the Pile validation set as the calibration data. For GPTQ, we use WikiText-2 as the calibration dataset, with 128 sequences of length 2048. All quantized models use a group size of 128 for weight quantization.

\textbf{Implementation Details.} Unlike the 65,536-token generation budget specified in the main text, Qwen3-4B is subject to a total context limit of 40,960 tokens.

\section{Additional Method Details}
\label{app:method_details}

\subsection{Candidate-Set Construction and QSTI Criteria}
\label{app:qsti_details}
\textbf{Candidate-Set Construction.} At each reasoning position $i$, we first construct a raw candidate pool from the union of the top-$p$ candidate sets produced by the quantized model $Q$ and the full-precision model $F$:

\begin{equation}
\mathcal{U}_{Q,i}
=
\mathcal{N}_{Q,i}\cup\mathcal{N}_{F,i},
\end{equation}

where $\mathcal{N}_{Q,i}$ and $\mathcal{N}_{F,i}$ denote the corresponding top-$p$ candidate sets under the shared prefix $x_{<t_i}$. We set $p=0.95$, consistent with the decoding configuration used in our other experiments. For tokens appearing in both candidate sets, we directly use the probabilities produced by the two models. For a token appearing in only one candidate set, we use a probability floor $\epsilon=10^{-3}$ for the missing side.

A probability floor may introduce an unreliable shift when it is larger than the observed probability on the available side. We therefore apply a filtering criterion to single-sided candidate events. Specifically, the retained candidate set is defined as

\begin{equation}
\mathcal{C}_{Q,i}
=
(\mathcal{N}_{Q,i}\cap\mathcal{N}_{F,i})
\cup\mathcal{S}_{Q,i}\cup\mathcal{S}_{F,i},
\end{equation}

where

\begin{equation}
\mathcal{S}_{Q,i}
=
\left\{k\in\mathcal{N}_{Q,i}\setminus\mathcal{N}_{F,i}:
p_Q(k|x_{<t_i})>\epsilon\right\},
\end{equation}

and

\begin{equation}
\mathcal{S}_{F,i}
=
\left\{k\in\mathcal{N}_{F,i}\setminus\mathcal{N}_{Q,i}:
p_F(k|x_{<t_i})>\epsilon\right\}.
\end{equation}

For each retained token $k\in\mathcal{C}_{Q,i}$, we compute the token-level probability shift as

\begin{equation}
\delta_{Q,i}(k)
=
\log p_Q(k|x_{<t_i})
-
\log p_F(k|x_{<t_i}).
\end{equation}

For a retained single-sided event, the probability on the missing side is set to $\epsilon$ only when computing this shift. For example, consider a token that appears only in the quantized candidate set and satisfies

\begin{equation}
p_Q(k|x_{<t_i})=0.003>\epsilon.
\end{equation}

Using $p_F(k|x_{<t_i})=\epsilon=0.001$ for shift estimation gives

\begin{equation}
\delta_{Q,i}(k)
=
\log 0.003-\log 0.001>0.
\end{equation}

In contrast, if the observed probability is $10^{-5}\leq\epsilon$, the event is excluded before shift estimation. In this case, assigning $\epsilon$ to the missing side would make the direction of the resulting shift primarily determined by the artificial floor rather than reliable model evidence. Only events retained in $\mathcal{C}_{Q,i}$ are included in the occurrence-level aggregation and subsequent QSTI statistics.

\textbf{Support and Stability Criteria.} After obtaining valid occurrence-level shifts, we further apply support and stability criteria to identify reliable quantization-sensitive tokens. For each token $k$, we denote $N_Q(k)$ as the number of valid shift events under quantizer $Q$, $S_Q(k)$ as the number of reference trajectories containing these events, and $\bar{\delta}_Q(k)$ as the average occurrence-level shift.

We first retain tokens that are observed under both AWQ and GPTQ and satisfy the token-cleaning rule. The cleaning rule removes special tokens, punctuation, non-English pieces, and uninterpretable tokenizer fragments. We then require sufficient event support from both quantizers:

\begin{equation}
N_{\mathrm{AWQ}}(k)>100,
\qquad
N_{\mathrm{GPTQ}}(k)>100.
\end{equation}

To ensure that the identified preference changes are not specific to a single quantization method, we further require the average shifts of AWQ and GPTQ to have the same direction:

\begin{equation}
\bar{\delta}_{\mathrm{AWQ}}(k)
\cdot
\bar{\delta}_{\mathrm{GPTQ}}(k)>0.
\end{equation}

Finally, we require sufficient trajectory coverage:

\begin{equation}
S_{\mathrm{AWQ}}(k)>105,
\qquad
S_{\mathrm{GPTQ}}(k)>105.
\end{equation}

The tokens satisfying these requirements form the shared candidate pool $K_{\mathrm{base}}$.

We then perform stability filtering on the positive-shift candidates in $K_{\mathrm{base}}$. Since RATIO aims to suppress tokens whose preference is increased by quantization, negative-shift tokens are not included in the final calibration set. For each positive token, we compute three complementary statistics. First, the conservative shift magnitude is defined as:

\begin{equation}
R_{\Delta}(k)
=
\min
\left(
|\bar{\delta}_{\mathrm{AWQ}}(k)|,
|\bar{\delta}_{\mathrm{GPTQ}}(k)|
\right).
\end{equation}

Second, we measure directional consistency across occurrence-level events:

\begin{equation}
R_{\mathrm{dir}}^{+}(k)
=
\min
\left(
\Pr\nolimits_{\mathrm{AWQ}}(\delta_i>0),
\Pr\nolimits_{\mathrm{GPTQ}}(\delta_i>0)
\right).
\end{equation}

Here, $\Pr\nolimits_{\mathrm{AWQ}}(\delta_i>0)$ represents the proportion of shift events where AWQ assigns a higher probability to token $k$ than the full-precision model. Third, we measure the proportion of events where the token appears in both the quantized and full-precision top-$p$ candidate sets:

\begin{equation}
R_{\mathrm{both}}(k)
=
\min
\left(
\Pr\nolimits_{\mathrm{AWQ}}(k\in \mathcal{N}_{Q,i}\cap\mathcal{N}_{F,i}),
\Pr\nolimits_{\mathrm{GPTQ}}(k\in \mathcal{N}_{Q,i}\cap\mathcal{N}_{F,i})
\right).
\end{equation}

These three statistics characterize the strength, consistency, and reliability of the observed quantization-related shifts. The corresponding thresholds are determined adaptively from the lower quartiles of the positive candidate distribution:

\begin{equation}
\tau_{\Delta}^{+}=\operatorname{Quantile}_{0.25}(R_{\Delta}),
\end{equation}

\begin{equation}
\tau_{\mathrm{dir}}^{+}=\operatorname{Quantile}_{0.25}(R_{\mathrm{dir}}^{+}),
\end{equation}

\begin{equation}
\tau_{\mathrm{both}}^{+}=\operatorname{Quantile}_{0.25}(R_{\mathrm{both}}).
\end{equation}

A token is retained as a numerically stable candidate when:

\begin{equation}
R_{\Delta}(k)\geq\tau_{\Delta}^{+},
\end{equation}

\begin{equation}
R_{\mathrm{dir}}^{+}(k)\geq\tau_{\mathrm{dir}}^{+},
\end{equation}

\begin{equation}
R_{\mathrm{both}}(k)\geq\tau_{\mathrm{both}}^{+}.
\end{equation}

The resulting set is denoted as $K_{\mathrm{num}}$.

To improve recall, we additionally consider a small reasoning-state lexical recovery branch within $K_{\mathrm{base}}$. This branch targets tokens with strong associations with reasoning-related behaviors but weaker numerical stability. Such tokens are not directly included in the calibration set; instead, they are passed to subsequent reasoning-context-aware validation for further verification. The resulting set is denoted as $K_{\mathrm{lex}}$.

The final QSTI candidate pool is defined as:

\begin{equation}
K_{\mathrm{QSTI}}
=
K_{\mathrm{num}}
\cup
K_{\mathrm{lex}}.
\end{equation}

All tokens in $K_{\mathrm{QSTI}}$ are further validated by RTV before being used for token-specific calibration.
 
\subsection{Reasoning-context-aware Token Validation}
\label{app:rtv_details}
\subsubsection{Numerical Evidence Construction} Given an input $x_r$, each quantized model $Q\in\{\mathrm{AWQ},\mathrm{GPTQ}\}$ freely generates a reasoning trajectory $y_r$. The full-precision model $F$ then evaluates the same quantized prefixes through teacher forcing. This allows all token-level comparisons to be performed under identical reasoning states. For each candidate $k\in K_{\mathrm{QSTI}}$, we construct the following quantitative evidence.

\textbf{Actual-token Scoring.} Suppose that the quantized model generates token $y_{r,t}=k$ at position $t$ in trajectory $r$. We measure its probability shift under the same quantized prefix as

\begin{equation}
\delta^{\mathrm{score}}_{Q,r,t}(k)
=
\log p_Q(k|y_{r,<t},x_r)
-
\log p_F(k|y_{r,<t},x_r).
\end{equation}

A positive value indicates that the quantized model assigns a higher probability to the generated token than the full-precision model under the same reasoning state. Let

\begin{equation}
\mathcal{O}_Q(k)
=
\{(r,t):y_{r,t}=k\}
\end{equation}

denote the set of actual-token events for $k$. We compute its event-weighted mean anomaly as

\begin{equation}
A_Q(k)
=
\frac{1}{|\mathcal{O}_Q(k)|}
\sum_{(r,t)\in\mathcal{O}_Q(k)}
\delta^{\mathrm{score}}_{Q,r,t}(k).
\end{equation}

Every occurrence receives equal weight in $A_Q(k)$. Consequently, trajectories in which the token occurs frequently contribute more strongly to this statistic.

\textbf{Trajectory-level Aggregation.} To measure whether the observed shift is consistent across different reasoning trajectories, let

\begin{equation}
I_Q(k)
=
\{r:\exists t,\,(r,t)\in\mathcal{O}_Q(k)\}
\end{equation}

denote the set of trajectories containing at least one actual-token event for $k$. Its trajectory support is defined as

\begin{equation}
S_Q(k)=|I_Q(k)|.
\end{equation}

Each trajectory contributes at most one unit to $S_Q(k)$, regardless of how frequently the token occurs within that trajectory. Let $n_{Q,r}(k)$ denote the number of occurrences of $k$ in trajectory $r$. We first compute its within-trajectory mean anomaly:

\begin{equation}
\bar{a}_{Q,r}(k)
=
\frac{1}{n_{Q,r}(k)}
\sum_{t:y_{r,t}=k}
\delta^{\mathrm{score}}_{Q,r,t}(k),
\end{equation}

and then average equally across trajectories:

\begin{equation}
T_Q(k)
=
\frac{1}{S_Q(k)}
\sum_{r\in I_Q(k)}
\bar{a}_{Q,r}(k).
\end{equation}

Unlike the event-weighted statistic $A_Q(k)$, $T_Q(k)$ assigns equal weight to each trajectory and therefore reduces the influence of individual long or repetitive responses. We define the corresponding cross-quantizer evidence as

\begin{equation}
E_{\mathrm{trajectory}}(k)
=
[T_{\mathrm{AWQ}}(k)>0]
\land
[T_{\mathrm{GPTQ}}(k)>0].
\end{equation}

This condition indicates that the positive anomaly persists across reasoning trajectories under both quantizers.

\textbf{Next-token Preview.} Actual-token scoring considers only tokens selected during generation. To capture tokens whose probabilities are increased by quantization even when they are not sampled, we additionally compare the top-$p$ candidate distributions under the same quantized prefix. For a token $k$ appearing in the quantized top-$p$ set, its preview shift is

\begin{equation}
\delta^{\mathrm{preview}}_{Q,r,t}(k)
=
\log p_Q(k|y_{r,<t},x_r)
-
\log p_F(k|y_{r,<t},x_r).
\end{equation}

If $k$ is absent from the full-precision top-$p$ set, we use the same probability floor as in QSTI:

\begin{equation}
p_{\mathrm{floor}}=0.001.
\end{equation}

This value is used only to approximate the missing-side probability and is unrelated to the inference-time penalty. We aggregate preview events similarly to actual-token events, obtaining trajectory support $S_Q^{\mathrm{preview}}(k)$ and mean anomaly $A_Q^{\mathrm{preview}}(k)$. Preview evidence is activated when

\begin{equation}
E_{\mathrm{preview}}(k)
=
\bigwedge_{Q\in\{\mathrm{AWQ},\mathrm{GPTQ}\}}
\left(
[S_Q^{\mathrm{preview}}(k)\geq20]
\land
[A_Q^{\mathrm{preview}}(k)>0]
\right).
\end{equation}

\textbf{Correctness-conditioned Evidence.} We use independent full-precision rollouts to divide the quantized trajectories into two primary groups:

\begin{equation}
\mathcal{E}_Q
=
\{r:Q\ \text{is incorrect and}\ F\ \text{is correct}\},
\end{equation}

\begin{equation}
\mathcal{C}_Q
=
\{r:Q\ \text{is correct and}\ F\ \text{is correct}\}.
\end{equation}

The first group represents reasoning failures introduced by quantization, while the second serves as a control group. For evidence type $u\in\{\mathrm{score},\mathrm{preview}\}$, let $\bar{a}_{Q,r}^{u}(k)$ denote the corresponding trajectory-level mean anomaly. We define the correctness-conditioned difference as

\begin{equation}
\Delta_Q^{u}(k)
=
\operatorname{mean}_{r\in\mathcal{E}_Q}\bar{a}_{Q,r}^{u}(k)
-
\operatorname{mean}_{r\in\mathcal{C}_Q}\bar{a}_{Q,r}^{u}(k).
\end{equation}

The means are taken over trajectories in which the corresponding event for $k$ is observed. A positive value indicates that the token exhibits a stronger quantization-related anomaly when the quantized model fails on a problem solved correctly by the full-precision model. We define

\begin{equation}
E_{\mathrm{score\mbox{-}correct}}(k)
=
\bigwedge_{Q\in\{\mathrm{AWQ},\mathrm{GPTQ}\}}
[\Delta_Q^{\mathrm{score}}(k)>0],
\end{equation}

and

\begin{equation}
E_{\mathrm{preview\mbox{-}correct}}(k)
=
\bigwedge_{Q\in\{\mathrm{AWQ},\mathrm{GPTQ}\}}
[\Delta_Q^{\mathrm{preview}}(k)>0].
\end{equation}

The two indicators are combined as

\begin{equation}
E_{\mathrm{correct}}(k)
=
E_{\mathrm{score\mbox{-}correct}}(k)
\lor
E_{\mathrm{preview\mbox{-}correct}}(k).
\end{equation}

Thus, $E_{\mathrm{correct}}(k)$ indicates that either actual-token or preview anomalies are consistently stronger in quantized-model failures than in shared successes under both quantizers.

\textbf{Loop and Repetition Evidence.}
We construct two complementary forms of repetition evidence: token enrichment in explicit-loop trajectories and localized high-repeat patterns.

\emph{Explicit-loop detection.}
We first label a trajectory as an explicit loop if it satisfies

\begin{equation}
\left(
\mathrm{repeat4}\geq80\%
\land
\mathrm{duplicate\mbox{-}lines}\geq50\%
\right)
\lor
\left(
\mathrm{max\ same\ token\ run}\geq64
\right).
\end{equation}

Here, we enumerate all contiguous 4-token windows in the trajectory. A window is treated as repeated if its exact 4-token sequence occurs at least twice within that trajectory, and all occurrences of such a sequence, including its first occurrence, are counted as repeated windows. The value of $\mathrm{repeat4}$ is the number of repeated-window occurrences divided by the total number of 4-token windows. Thus, $\mathrm{repeat4}\geq80\%$ means that at least 80\% of the trajectory's 4-token windows have an identical counterpart elsewhere in the same trajectory. To compute $\mathrm{duplicate\mbox{-}lines}$, we split the decoded trajectory at newline characters and discard empty lines and lines shorter than a predefined length threshold. We then compare the remaining lines using exact string matching. A retained line is counted as duplicated if an identical line appears elsewhere in the same trajectory, and $\mathrm{duplicate\mbox{-}lines}$ is the proportion of such duplicated line occurrences among all retained lines. Finally, $\mathrm{max\ same\ token\ run}$ denotes the longest consecutive repetition of a single token. An explicit loop that produces an incorrect answer is assigned to the error-loop set $L_Q^{\mathrm{error}}$, whereas a trajectory that does not satisfy the loop criterion is assigned to the non-loop set $L_Q^{\mathrm{nonloop}}$.

\emph{Token enrichment in explicit loops.}
After labeling the trajectories, we determine whether candidate token $k$ occurs more frequently in error-loop trajectories than in non-loop trajectories. For evidence type $u\in\{\mathrm{score},\mathrm{preview}\}$, its length-normalized occurrence rate in trajectory $r$ is

\begin{equation}
f_{Q,r}^{u}(k)
=
1000
\frac{\operatorname{count}_{Q,r}^{u}(k)}
{\operatorname{length}(y_r)}.
\end{equation}

We then compute the loop-enrichment ratio

\begin{equation}
R_Q^{\mathrm{loop},u}(k)
=
\frac{
\operatorname{mean}_{r\in L_Q^{\mathrm{error}}}
f_{Q,r}^{u}(k)
}{
\operatorname{mean}_{r\in L_Q^{\mathrm{nonloop}}}
f_{Q,r}^{u}(k)
}.
\end{equation}

The corresponding cross-quantizer indicators are

\begin{equation}
E_{\mathrm{score\mbox{-}loop}}(k)
=
\bigwedge_{Q\in\{\mathrm{AWQ},\mathrm{GPTQ}\}}
[R_Q^{\mathrm{loop},\mathrm{score}}(k)>1],
\end{equation}

\begin{equation}
E_{\mathrm{preview\mbox{-}loop}}(k)
=
\bigwedge_{Q\in\{\mathrm{AWQ},\mathrm{GPTQ}\}}
[R_Q^{\mathrm{loop},\mathrm{preview}}(k)>1].
\end{equation}

If the mean occurrence rate in non-loop trajectories is zero while that in error-loop trajectories is positive, the ratio is treated as infinity. If both rates are zero, the ratio is treated as missing and does not activate the corresponding indicator.

\emph{Localized high-repeat evidence.}
The trajectory-level loop criterion may overlook localized repetition. We therefore inspect all 4-token windows covering each occurrence of $k$. An occurrence is marked as a high-repeat event if any such exact 4-token sequence occurs at least 8 times within the same trajectory. Let $H_Q(k)$ denote the number of trajectories containing at least one high-repeat event for $k$. We define

\begin{equation}
E_{\mathrm{repeat}}(k)
=
[H_{\mathrm{AWQ}}(k)>0]
\land
[H_{\mathrm{GPTQ}}(k)>0].
\end{equation}

Because this condition is intentionally permissive, it is used only as supporting evidence.

\textbf{Evidence Consolidation.} Finally, we consolidate the raw indicators into four distinct evidence categories. We retain $E_{\mathrm{trajectory}}(k)$ and $E_{\mathrm{preview}}(k)$ as separate categories, and define

\begin{equation}
E_{\mathrm{loop/repeat}}(k)
=
E_{\mathrm{score\mbox{-}loop}}(k)
\lor
E_{\mathrm{preview\mbox{-}loop}}(k)
\lor
E_{\mathrm{repeat}}(k).
\end{equation}

Together with $E_{\mathrm{correct}}(k)$ defined above, the number of distinct evidence categories is

\begin{equation}
N_C(k)
=
E_{\mathrm{trajectory}}(k)
+
E_{\mathrm{preview}}(k)
+
E_{\mathrm{correct}}(k)
+
E_{\mathrm{loop/repeat}}(k).
\end{equation}

Each category contributes at most one count, preventing closely related indicators from being counted repeatedly.

\subsubsection{Staged Token Selection}

Based on the evidence constructed above, RTV progressively selects tokens through a strict core and two expansion stages. All stages operate exclusively on $K_{\mathrm{QSTI}}$, ensuring that contextual validation cannot introduce tokens that have not passed QSTI.

\textbf{Strict core.}
We first construct a reference pool containing candidates with sufficient trajectory support and positive actual-token anomalies under both quantizers:

\begin{equation}
K_{\mathrm{ref}}
=
\left\{
k\in K_{\mathrm{QSTI}}:
\bigwedge_{Q\in\{\mathrm{AWQ},\mathrm{GPTQ}\}}
\left[
S_Q(k)\geq20
\land
A_Q(k)>0
\right]
\right\}.
\end{equation}

For each quantizer, we compute the lower quartile of its anomaly values within the reference pool:

\begin{equation}
\tau_Q
=
\operatorname{Quantile}_{0.25}
\left(
\left\{
A_Q(j):j\in K_{\mathrm{ref}}
\right\}
\right),
\qquad
Q\in\{\mathrm{AWQ},\mathrm{GPTQ}\}.
\end{equation}

A candidate passes the strict numerical gate if

\begin{equation}
G_{\mathrm{strict}}(k)
=
[k\in K_{\mathrm{QSTI}}]
\land
\bigwedge_{Q\in\{\mathrm{AWQ},\mathrm{GPTQ}\}}
\left[
S_Q(k)\geq20
\land
A_Q(k)\geq\tau_Q
\right].
\end{equation}

The thresholds $\tau_Q$ are derived separately for each model and quantizer rather than shared across models. This gate therefore requires substantial positive anomalies under both AWQ and GPTQ, rather than deviations that are only marginally above zero. Candidates passing the gate are further examined using reasoning-state lexical information. A token is retained in the strict core only when its observed use is directly associated with reasoning operations such as restarting, backtracking, verification, hesitation, self-questioning, self-correction, uncertainty, or metacognitive control.

\textbf{First-round expansion (R1).}
To recover candidates with moderate support or asymmetric anomaly strength across quantizers, R1 applies a relaxed numerical gate $G_{\mathrm{R1}}(k)$ defined by:

\begin{equation}
[k\in K_{\mathrm{QSTI}}]
\land
\bigwedge_{Q\in\{\mathrm{AWQ},\mathrm{GPTQ}\}}
[S_Q(k)\geq10]
\land
\bigvee_{Q\in\{\mathrm{AWQ},\mathrm{GPTQ}\}}
[A_Q(k)>0]
\land
[N_C(k)\geq1]
\land
\neg G_{\mathrm{strict}}(k).
\end{equation}

All R1 candidates are first examined using the same reasoning-state lexical information as the strict core. We additionally require contextual review when the candidate is supported by only one evidence category and that category is either trajectory-level or preview evidence. The contextual-review trigger is

\begin{equation}
[N_C(k)=1]
\land
[E_{\mathrm{trajectory}}(k)\lor E_{\mathrm{preview}}(k)]
\land
\neg E_{\mathrm{correct}}(k)
\land
\neg E_{\mathrm{loop/repeat}}(k).
\end{equation}

This condition is deterministic. It is activated because trajectory-level and preview evidence primarily reflect distributional deviations; without correctness-conditioned or repetition-related support, the surrounding reasoning context must be inspected to distinguish inefficient reasoning from valid verification or self-correction.

\textbf{Second-round expansion (R2).}
R2 considers near-miss candidates with weaker numerical support but evidence from multiple complementary categories:

\begin{equation}
G_{\mathrm{R2}}(k)
=
[k\in K_{\mathrm{QSTI}}]
\land
\bigwedge_{Q\in\{\mathrm{AWQ},\mathrm{GPTQ}\}}
[S_Q(k)\geq5]
\land
[N_C(k)\geq2]
\land
\neg G_{\mathrm{strict}}(k)
\land
\neg G_{\mathrm{R1}}(k).
\end{equation}

Because R2 uses the most relaxed support threshold, every candidate passing this gate undergoes contextual review. We inspect its occurrences in the original quantized trajectories to determine whether the token is consistently associated with inefficient deliberation, rather than necessary verification, correction, or other valid reasoning behavior.

Let $K_{\mathrm{strict}}$, $K_{\mathrm{R1}}$, and $K_{\mathrm{R2}}$ denote the tokens retained by the three stages. The final model-specific token set is

\begin{equation}
K
=
K_{\mathrm{strict}}
\cup
K_{\mathrm{R1}}
\cup
K_{\mathrm{R2}}.
\end{equation}

Contextual review is applied only after the corresponding numerical gate has been satisfied. It determines whether a numerically qualified candidate is retained or excluded, but never introduces tokens outside the gated candidate pools.

\subsection{Derivation of Token-Specific Logit Correction}
\label{app:tspd_derivation}
For a target token $k$, its probability under the softmax function can be written as:

\begin{equation}
p(k)=
\frac{e^{z_k}}
{e^{z_k}+\sum_{j\neq k}e^{z_j}},
\end{equation}

where $z_k$ denotes the logit of token $k$. Let

\begin{equation}
S=\sum_{j\neq k}e^{z_j},
\end{equation}

then the probability of token $k$ can be rewritten as:

\begin{equation}
p(k)=
\frac{e^{z_k}}
{e^{z_k}+S}.
\end{equation}

The corresponding logit transformation is defined as:

\begin{equation}
\operatorname{logit}(p(k))
=
\log\frac{p(k)}{1-p(k)}.
\end{equation}

Substituting the probability expression into the logit transformation gives:

\begin{equation}
\operatorname{logit}(p(k))=\log\frac{\frac{e^{z_k}}{e^{z_k}+S}}{\frac{S}{e^{z_k}+S}}=\log\frac{e^{z_k}}{S}=z_k-\log S.
\end{equation}

Under the independent token calibration assumption, only the target token logit is adjusted while all other quantized logits remain unchanged. Let $z'_{Q,k}$ denote the corrected logit whose probability matches the full-precision preference. The required correction magnitude is therefore:

\begin{equation}
c_{Q,i}(k)
=z_{Q,k}-z'_{Q,k}
=\operatorname{logit}(p_Q(k|x_{<t_i}))
-\operatorname{logit}(p_F(k|x_{<t_i})).
\end{equation}

\section{Dialogue Examples}
\label{app:dialogue_examples}

Table~\ref{tab:conversation-analysis} shows how RATIO curbs repetition and erroneous continuations while preserving correct answers.

\begin{table}[H]
\centering
\caption{Dialogue examples on Qwen-1.5B with GPTQ-W3. Red marks repetitive or erroneous reasoning; green marks correct RATIO calculations and answers. Ellipses indicate omissions.}
\label{tab:conversation-analysis}
\setlength{\tabcolsep}{5pt}
\renewcommand{\arraystretch}{1.15}
\small
\begin{tabular}{@{}p{0.12\textwidth}p{0.84\textwidth}@{}}
\toprule
\multicolumn{2}{c}{\textbf{Case 1: Repetitive reasoning after reaching the correct result}} \\
\midrule
\textbf{Prompt} & From her science class study, Brady learned that each whale has 40 gallons of blood and that a shark has three times as much blood as a whale. How many gallons of blood do ten sharks have? \\
\textbf{GPTQ-W3} & Each shark has $3\times40=120$ gallons, so ten sharks have $10\times120=1200$ gallons. Wait, let me double-check that. $\ldots$ \textcolor{red}{Wait, that seems correct. Let me think if I interpreted the question correctly.} $\ldots$ \textcolor{red}{Alternatively, perhaps I can think of the problem as ten sharks each having three times as much as a whale, so $10\times(3\times W)=3\times(10\times W)$. Therefore, ten sharks have $3\times10\times40=1200$ gallons. Alternatively, perhaps I can think of it as $10\times(3\times W)=3\times(10\times W)$. $\ldots$ Wait, 3 W. $\ldots$ 3 W3 W3 W.} \\
\textbf{RATIO} & \textcolor{green!50!black}{A shark has three times as much blood as a whale. Therefore, each shark has $3\times40=120$ gallons of blood. Ten such sharks have $10\times120=1200$ gallons of blood. The final answer is $\boxed{1200}$.} \\
\midrule
\multicolumn{2}{c}{\textbf{Case 2: Repetition followed by an incorrect reasoning trajectory}} \\
\midrule
\textbf{Prompt} & Compute $99^2+99+1$ in your head. \\
\textbf{GPTQ-W3} & First, $99^2=(100-1)^2=10{,}000-200+1=9{,}801$. Adding 99 gives $9{,}900$, and adding 1 gives $9{,}901$. $\ldots$ Let me verify each step again. $99^2$ is 9,801, then $9{,}801+99=9{,}900$, and adding 1 gives 9,901. $\ldots$ \textcolor{red}{Alternatively, maybe I can factor it as $99\times(99+1)+1$. That gives $99\times100+1=9{,}901$. Alternatively, if I think of $99^2+99+1$ as $(99+1)^2-1$, that is $100^2-1=9{,}999$. Wait, no, that is incorrect. $\ldots$ So, 9999. $\ldots$ 1099999999999999999999999999999999999999.} \\
\textbf{RATIO} & \textcolor{green!50!black}{First, $99^2=(100-1)^2=10{,}000-200+1=9{,}801$. Next, $9{,}801+99=9{,}900$, and $9{,}900+1=9{,}901$. Thus, the final answer is $\boxed{9901}$.} \\
\bottomrule
\end{tabular}
\end{table}

\end{document}